\documentclass{article}

\PassOptionsToPackage{numbers,sort&compress}{natbib}
\usepackage[final]{neurips_2022}

\usepackage{chngcntr}
\usepackage{enumitem}
\usepackage{multirow}
\usepackage{graphicx}
\usepackage[utf8]{inputenc} 
\usepackage[T1]{fontenc}    
\usepackage{hyperref}       
\usepackage{url}            
\usepackage{booktabs}       
\usepackage{amsfonts}       
\usepackage{nicefrac}       
\usepackage{microtype}      
\usepackage{xcolor}         
\usepackage{amsmath}

\DeclareMathOperator*{\argmin}{arg\,min}
\usepackage{mathtools}

\usepackage{enumitem}

\usepackage[acronym,shortcuts]{glossaries}
\newacronym{CNN}{CNN}{convolutional neural network}
\newacronym{DNN}{DNN}{deep neural network}
\newacronym{MAE}{MAE}{mean absolute error}
\newacronym{MSE}{MSE}{mean squared error}
\newacronym{NFB}{NFB}{nerve fiber bundle}
\newacronym{HNA}{HNA}{hybrid neural autoencoder}
\newacronym{SPV}{SPV}{simulated prosthetic vision}
\newacronym{FC}{FC}{fully connected}
\newacronym{BN}{BN}{batch normalization}
\newacronym{RA}{RA}{relative accuracy}
\newacronym{VPU}{VPU}{vision processing unit}
\glsdisablehyper  

\title{Hybrid Neural Autoencoders for Stimulus Encoding in Visual and Other Sensory Neuroprostheses}

\author{%
  Jacob Granley \\
  Department of Computer Science\\
  University of California, Santa Barbara\\
  \texttt{jgranley@ucsb.edu} \\
  \And
   Lucas Relic \\
   Department of Computer Science \\
   University of California, Santa Barbara \\
   \texttt{lucasrelic@ucsb.edu} \\
   \And
   Michael Beyeler \\
   Department of Computer Science \\
   Department of Psychological \& Brain Sciences \\
   University of California, Santa Barbara \\
   \texttt{mbeyeler@ucsb.edu} \\
}

\begin{document}

\maketitle

\begin{abstract}
Sensory neuroprostheses are emerging as a promising technology to restore lost sensory function or augment human capabilities.
However, sensations elicited by current devices often appear artificial and distorted.
Although current models can predict the neural or perceptual response to an electrical stimulus, an optimal stimulation strategy solves the inverse problem: what is the required stimulus to produce a desired response?
Here, we frame this as an end-to-end optimization problem, where a deep neural network stimulus encoder is trained to invert a known and fixed forward model that approximates the underlying biological system.
As a proof of concept, we demonstrate the effectiveness of this \acf{HNA} in visual neuroprostheses.
We find that \acs{HNA} produces high-fidelity patient-specific stimuli representing handwritten digits and segmented images of everyday objects, and significantly outperforms conventional encoding strategies across all simulated patients.
Overall this is an important step towards the long-standing challenge of restoring high-quality vision to people living with incurable blindness and may prove a promising solution for a variety of neuroprosthetic technologies.
\end{abstract}

\section{Introduction}

Sensory neuroprostheses are emerging as a promising technology to restore lost sensory function or augment human capacities \cite{cinel_neurotechnologies_2019,fernandez_development_2018}.
In such devices, diminished sensory modalities (e.g., hearing \cite{wilson_better_1991}, vision \cite{fernandez_visual_2021,luo_argus_2016}, cutaneous touch \cite{tan_neural_2014}) are re-enacted through streams of artificial input to the nervous system.
For example, visual neuroprostheses electrically stimulate neurons in the early visual system to elicit neuronal responses that the brain interprets as visual percepts.
Such devices have the potential to restore a rudimentary form of vision to millions of people living with incurable blindness.

\begin{figure}[t]
    \centering
    \includegraphics[width=\linewidth]{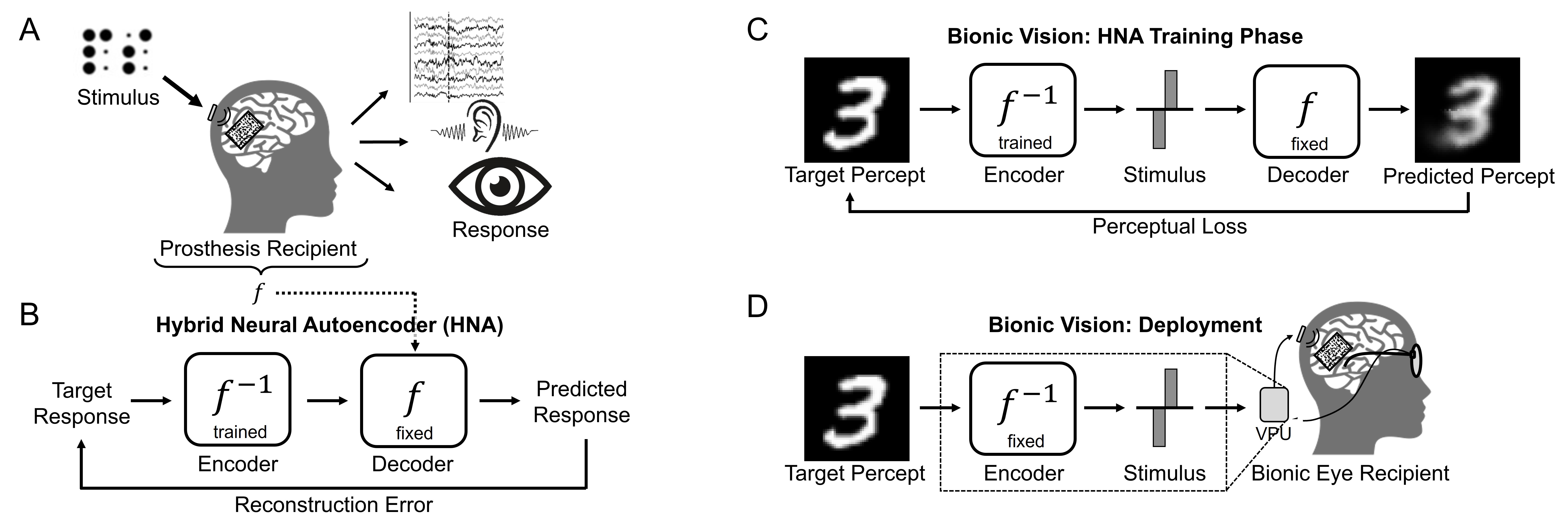}
    \caption{\emph{A)} Sensory neuroprosthesis. A forward model ($f$) is used to approximate the neuronal or, ideally, perceptual response to electrical stimuli.
    \emph{B)} \Acf{HNA}. A deep neural encoder ($f^{-1}$) is trained to predict the patterns of electrical stimulation that elicit responses closest to the target.
    \emph{C)} Visual neuroprostheses are one prominent application of \acs{HNA}, where an encoder can be trained to predict the electrical stimulation needed to elicit a desired visual percept.
    \emph{D)} The trained encoder is deployed on a \acf{VPU}, predicting stimuli in real-time that are decoded by the patient's visual cortex.}
    \label{fig:overview}
\end{figure}

However, all of these technologies face the challenge of interfacing with a highly nonlinear system of biological neurons whose role in perception is not fully understood. Due to the limited resolution of electrical stimulation, prostheses often create neural response patterns foreign to the brain. 
Consequently, sensations elicited by current sensory neuroprostheses often appear artificial and distorted \cite{erickson-davis_what_2021,murray_embodiment_2008}.
A major outstanding challenge is thus to identify a stimulus encoding that leads to perceptually intelligible sensations.
Often there exists a forward model, $f$ (Fig.~\ref{fig:overview}A), constrained by empirical data, that can predict a neuronal or (ideally) perceptual response to the applied stimulus (see \cite{brunton_data-driven_2019} for a recent review).
To find the stimulus that will elicit a desired response, one essentially needs to find the inverse of the forward model, $f^{-1}$. However, realistic forward models are rarely analytically invertible, making this a challenging open problem for neuroprostheses.

Here we propose to approach this as an end-to-end optimization problem, where a \ac{DNN} (\emph{encoder}) is trained to invert a known, fixed forward model (\emph{decoder}, Fig.~\ref{fig:overview}B).
The encoder is trained to predict the patterns of electrical stimulation patterns that elicit perception (\emph{e.g.,} vision, audition) or neural responses (\emph{e.g.,} firing rates) closest to the target. This \ac{HNA} could in theory be used to optimize stimuli for any open-loop neuroprosthesis with a known forward model that approximates the underlying biological system.

In order to optimize end-to-end, the forward model must be differentiable and computationally efficient.
When this is not the case, an alternative approach is to train a surrogate neural network, $\hat{f}$, to approximate the forward model \cite{relic_deep_2022, montes_de_oca_zapiain_accelerating_2021, nabian_deep_2019, nikolopoulos_non-intrusive_2022}. However, even well-trained surrogates may result in errors when included in our end-to-end framework, due to the encoders' ability to learn to exploit the surrogate model. We therefore also evaluate whether a surrogate approach to \ac{HNA} is suitable. 

To this end, we make the following contributions:
\begin{itemize}[topsep=0pt,leftmargin=15pt,parsep=0pt]
    \item We propose a \acf{HNA} consisting of a deep neural encoder trained to invert a fixed, numerical or symbolic forward model (decoder) to optimize stimulus selection. Our framework is general and addresses an important challenge with sensory neuroprostheses. 
    
    \item As a proof of concept, we demonstrate the utility of \ac{HNA} for visual neuroprostheses, where we predict electrode activation patterns required to produce a desired visual percept. 
    We show that the \acs{HNA} is able to produce high-fidelity, patient-specific stimuli representing handwritten digits and segmented images of everyday objects, drastically outperforming conventional approaches. 
    \item We evaluate replacing a computationally expensive or nondifferentiable forward model with a surrogate,
    highlighting benefits and potential dangers of popular surrogate techniques.
\end{itemize}

\section{Background}

\paragraph{Sensory Neuroprostheses} 
Recent advances in neural understanding, wearable electronics, and biocompatible materials have accelerated the development of sensory neuroprostheses to restore perceptual function to people with impaired sensation.
Use of neuroprostheses typically requires invasive implants that elicit neural responses via electrical, magnetic, or optogenetic stimulation. Two of the most promising applications are cochlear implants, which stimulate the auditory nerve to elicit sounds \cite{wilson_better_1991}, and visual implants (see next subsection) to restore vision to the blind. However, a variety of other devices are in development; for instance, to restore touch \citep{tan_neural_2014, tabot_restoring_2013} or motor function \citep{capogrosso_brainspine_2016}.
The latter differ from other sensory neuroprostheses in that they generate stimuli using motor feedback (\emph{closed loop}) \citep{chapman_multifunctional_2018, wagner_targeted_2018}.
In the absence of feedback (\emph{open loop}), a proper stimulus encoding is paramount to the success of these devices.

\begin{figure}[t]
    \centering
    \includegraphics[width=.9\linewidth]{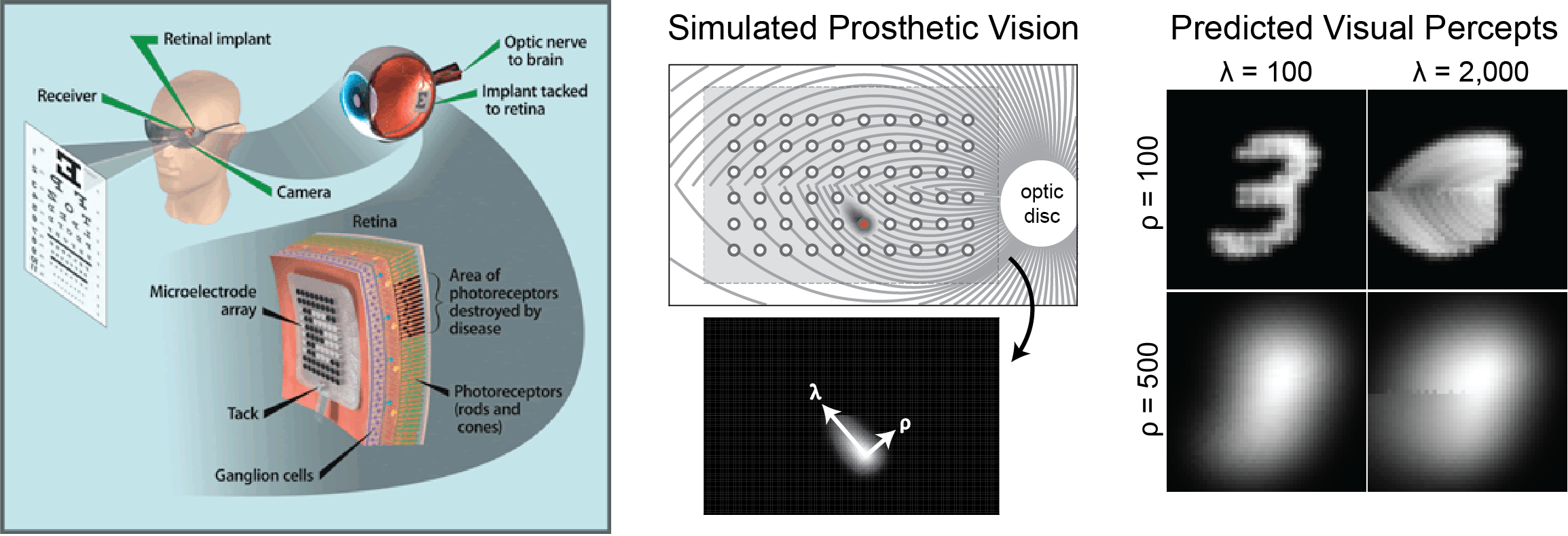}
    \caption{
        \emph{Left}: Visual prosthesis. Incoming target images are transmitted from a camera to an implant in the retina, which encodes the image as an electrical stimulus pattern. \emph{Center}: Electrical stimulation (red disc) of a nerve fiber bundle (gray lines) leads to elongated tissue activation (gray shaded region) and phosphenes whose shape can be described by two parameters, $\lambda$ (axonal spread) and $\rho$ (radial spread).
        \emph{Right}: Predicted percepts for an MNIST digit using varying $\rho$ and $\lambda$ values.
        }
    \label{fig:axonmapmodel}
\end{figure}

\paragraph{Restoring Vision to the Blind}
For millions of people who are living with incurable blindness, a visual prostheses (\emph{bionic eye}, Fig.~\ref{fig:axonmapmodel}, \emph{left}) may be the only treatment option \cite{ayton_update_2020}.
Analogous to cochlear implants, these devices electrically stimulate surviving cells in the visual pathway to evoke visual percepts (\emph{phosphenes}), which can support simple behavioral tasks
\cite{luo_argus_2016,stingl_interim_2017,karapanos_functional_2021}.

A common misconception is that each electrode in the array can be thought of as a pixel in an image; to generate a complex visual experience, one then simply needs to turn on the right combination of pixels \cite{dobelle_artificial_2000}.
However, recent evidence suggests that phosphenes often appear distorted (\emph{e.g.}, as lines, wedges, and blobs) and vary drastically across subjects and electrodes \cite{erickson-davis_what_2021,fernandez_visual_2021}. 

Phosphene appearance has been best characterized in epiretinal implants, where inadvertent activation of \acp{NFB} in the optic fiber layer of the retina leads to elongated phosphenes \cite{beyeler_model_2019,rizzo_perceptual_2003} (Fig.~\ref{fig:axonmapmodel}, \emph{center}).
To this end, Granley \emph{et.~al} \citep{granley_computational_2021} developed a forward model to predict phosphene shape as a function of both neuroanatomical parameters (\emph{i.e.}, location of the stimulating electrode) and stimulus parameters (\emph{i.e.}, pulse frequency, amplitude, and duration).
%
With this model, phosphenes are primarily characterized by two main parameters, $\rho$ and $\lambda$, which dictate the size and elongation of the elicited phosphene, respectively (Fig.~\ref{fig:axonmapmodel}, \emph{right}). 
These parameters can be determined using psychophysical tasks (\emph{e.g.}, drawings, brightness ratings) \cite{beyeler_model_2019, granley_computational_2021}, and although they vary drastically across patients \cite{beyeler_model_2019}, they do not change much over time \cite{luo_long-term_2016,beyeler_learning_2017}.
Stimulation from multiple electrodes is nonlinearly integrated into a combined perception, and if two electrodes happen to activate the same \ac{NFB}, they might not generate two distinct phosphenes. 

\section{Related Work}
The conventional `naive' encoding strategy 
sets
the amplitude of each electrode to the brightness of the corresponding pixel in the target image \cite{chen_simulating_2009,luo_argus_2016}, making the stimulus a down-sampled version of the target.
Although simple, this strategy only works with near-linear forward models, cannot account for real phosphene data, and often leads to unrecognizable percepts (Fig.~\ref{fig:axonmapmodel}, \emph{right}) \cite{beyeler_model_2019,erickson-davis_what_2021}.

Many alternative stimulation strategies have been proposed~\cite{tong2020stimulation}. Shah \textit{et al.}~\cite{shah_optimization_2019} used a greedy approach to iteratively select the stimuli that best recreate a desired neural activity pattern over a given temporal window, assuming that the brain would integrate them into a coherent visual percept. Ghaffari \textit{et al.}~\citep{ghaffari_closed-loop_2021} used a neural network surrogate model combined with an interior point algorithm to optimize for localized, circular neural activation patterns for individual electrodes. Fauvel \textit{et al.}~\cite{fauvel_human---loop_2021} used human in-the-loop Bayesian optimization to achieve encodings perceptually favored by the simulated patient. 
Spencer \textit{et al.}~\cite{spencer_global_2019} proposed framing stimulus encoding as inversion of a forward model of neural activation patterns, but to approximate the inverse, their approach either requires simplification or is NP-hard \cite{spencer_global_2019}.

Van Steveninck \textit{et al.} \cite{de_ruyter_van_steveninck_end--end_2022} proposed an end-to-end optimization strategy with a fixed phosphene model, similar to \ac{HNA}. However, their approach crucially differs from ours in its inclusion of a secondary \ac{DNN} to post-process the predicted phosphenes. This is a critical limitation, because a low reconstruction loss does not reveal whether a high-fidelity encoder was learned or whether the secondary decoder network simply learned to correct for the encoder's mistakes.
In addition, they used an unrealistic phosphene model that simply upscales and smooths a binary stimulus pattern.  
It is therefore not clear whether their results would generalize to real visual prosthesis patients.

Relic \textit{et al.}~\cite{relic_deep_2022} also utilized the end-to-end approach, but without the secondary decoder network used in \cite{de_ruyter_van_steveninck_end--end_2022}. They used a more realistic phosphene model, which accounts for some spatial distortions (\textit{e.g.}, axonal streaks), but not the effects of stimulus parameters. Since including a realistic phosphene model in the loop is not straightforward, they instead trained a surrogate neural network to approximate the forward model. We re-implemented Relic's surrogate approach in this paper as a baseline method to compare against, as described in Section~\ref{sec:forward}.

Taken together, we identified three main limitations of previous work that this study aims to address:
\begin{itemize}[topsep=0pt,leftmargin=15pt,parsep=0pt]
    \item[1)] \textbf{Unrealistic forward models.} Most previous approaches (\emph{e.g.}, \cite{de_ruyter_van_steveninck_end--end_2022,shah_optimization_2019,spencer_global_2019}) use an overly simplified forward model that cannot account for empirical data \cite{beyeler_model_2019,erickson-davis_what_2021}.
    We overcome this limitation by developing (and inverting) a differentiable version of a neurophysiologically informed and psychophysically validated phosphene model \cite{granley_computational_2021} that can account for the effects of stimulus amplitude, frequency, and pulse duration on phosphene appearance. 
    \item[2)] \textbf{Optimization of neural responses.} 
    Most previous approaches (\emph{e.g.}, \cite{shah_optimization_2019,spencer_global_2019}) focus on optimizing neural activation patterns in the retina in response to electrical stimulation (``bottom-up'').
    However, the visual system undergoes extensive remodeling during  blinding diseases such as retinitis pigmentosa \cite{marc_remodeling_2003}.
    Thus the link between neural activity and visual perception is unclear.
    We overcome this limitation by inverting a phenomenological (``top-down'') model constrained by behavioral data that predicts visual perception directly from electrical stimuli \cite{beyeler_model_2019,granley_computational_2021}.
    \item[3)] \textbf{Objective function}
    Most previous approaches rely on minimizing \ac{MSE} between the target and reconstructed image.
    Although simple and efficient, \ac{MSE} is known to be a poor measure of perceptual dissimilarity for images \citep{wang_mean_2009} and does not align well with human assessments of image quality \citep{zhai_perceptual_2020}.
    We overcome this limitation by proposing a joint perceptual metric that combines \ac{MAE}, VGG, and Laplacian smoothing losses.
\end{itemize}

\section{Methods}

\paragraph{Problem Formulation} 
We consider a system where there is some known forward process $f$ mapping stimuli to responses $f : \mathcal{S} \mapsto \mathcal{R}, f(\mathcal{S}) \subset \mathcal{R}$. 
In the case of visual prostheses, $f$ may map stimuli to visual percepts. 
$f$ may optionally be parameterized by patient-specific parameters $\phi$.

Finding the best stimulus for an arbitrary target response $\mathbf{t} \in \mathcal{R}$ is equivalent to finding the inverse of $f$.
However, since not every response can be perfectly reproduced by a stimulus, the true inverse of $f$ is not well defined.
We therefore seek the pseudoinverse (still denoted as $f^{-1}$ for simplicity) instead, which outputs the stimuli that produce the closest response to $\mathbf{t}$ under some distance metric $d$:
\begin{equation}
    f^{-1}(\mathbf{t}, \phi) \coloneqq \underset{\mathbf{s}\in\mathcal{S}}{\argmin}\,d(f(\mathbf{s}; \phi), \mathbf{t}).
\end{equation} 

This problem is straightforward to solve using an autoencoder approach, where a learned encoder $f^{-1}$ is trained to invert the fixed decoder $f$ (i.e., forward model).

\paragraph{Encoder}
We approximate the pseudoinverse $f^{-1}$ with a \ac{DNN} encoder $\hat{f}^{-1}(\mathbf{t}, \phi; \theta)$ with weights $\theta$, trained to minimize the distance $d$ between the target image $\mathbf{t}$ and predicted image $\hat{\mathbf{t}}$:
\begin{equation}
   \underset{\theta,\,\phi \sim p(\phi)}{\min}\,\,  d(\mathbf{t},\hat{\mathbf{t}})
\end{equation}
\begin{equation}
    \hat{\mathbf{t}}  = f(\,\hat{f}^{-1}(\mathbf{t}, \phi;\, \theta);\,\, \phi),
\end{equation}
where $\phi$ is sampled from a uniform random distribution spanning the empirically observed range of patient-specific parameters \cite{beyeler_model_2019, granley_computational_2021}.

We use a simple architecture consisting solely of \ac{FC} and \ac{BN} \cite{ioffe_batch_2015} layers (1.4M trainable parameters). First, the target $\mathbf{t}$ is flattened and input to a \ac{FC} layer. In parallel, the patient parameters $\phi$ are input to a \ac{BN} layer and two hidden \ac{FC} layers. Then, the outputs of these two paths are concatenated, and the combined vector fed through two \ac{FC} layers, producing an intermediate representation $\mathbf{x}$. Amplitudes are predicted from $\mathbf{x}$ with a \ac{FC} layer. The amplitudes are then concatenated with $\mathbf{x}$, put through a \ac{BN} layer, and used to predict frequency and pulse duration, each with a \ac{FC} layer. 
The outputs are merged into a stimulus matrix $\mathbf{\hat{s}}$. 
All layers use leaky ReLU activation, except for stimulus outputs, which use ReLU to enforce nonnegativity. 

\paragraph{Decoder} \label{sec:forward}
The \ac{HNA} decoder is a differentiable approximation of the underlying biological system, and describes the transform from stimulus to response. For our decoder $f$, we use a reformulated but equivalent version of the model described in \cite{granley_computational_2021}. This model is specific to epiretinal prostheses; analogous models exist for other  neuroprostheses  (\textit{e.g.}, auditory \cite{dorman2005acoustic, svirsky2013validation, dorman2003simulations, dorman1997speech, cooper2008music, loizou2000speech}, tactile and somatosensory \cite{saal2017simulating, okorokova2018biomimetic, weber2012interfacing, kim2010predicting, mileusnic2006mathematical}), and could potentially be adapted for use with \ac{HNA}. We use a square $15 \times 15$ array of 150$\mu$m electrodes, spaced 400$\mu$m apart and centered on the fovea. The size and scale of this device are motivated by similar designs in real epiretinal implants.

$f$ takes as input a stimulus matrix $\mathbf{s} \in \mathbb{R}_{\geq0}^{n_e \mathrm{x} 3}$, where the stimulus on each electrode ($\mathbf{s_e}$) is a biphasic pulse train described by its frequency, amplitude, and pulse duration. A simulated map of retinal \ac{NFB}s is used to predict phosphene shape. Following \cite{beyeler_model_2019}, each retinal ganglion cells' activation is assumed to be the maximum stimulation intensity along its axon. Axon sensitivity is assumed to decay exponentially with i) distance to the stimulating electrode (radial decay rate, $\rho$) and distance to the soma along the curved axon (axonal decay rate, $\lambda$).
To account for stimulus properties \citep{granley_computational_2021}, $\rho$, $\lambda$, and the per-electrode brightness are scaled by three simple equations $F_\mathrm{size}(\mathbf{s_e}, \phi)$, $F_\mathrm{streak}(\mathbf{s_e}, \phi)$, and $F_\mathrm{bright}(\mathbf{s_e}, \phi)$, respectively. 

The brightness of a pixel located at the point $\mathbf{x} \in \mathbb{R}^2$ in the output image is given by 
\begin{equation} \label{eq:forward1}
    f(\mathbf{s}; \phi) = \max_{\mathbf{a} \in A}\sum_{e \in E}F_\mathrm{bright}(\mathbf{s_e}, \phi)  \exp\left(\frac{-|| \mathbf{x} - \mathbf{e}||_2^2}{2\rho^2 F_\mathrm{size}(\mathbf{s_e}, \phi) } + \frac{-d_{s}(\mathbf{x}, \mathbf{a})^2}{2\lambda^2 F_\mathrm{streak}(\mathbf{s_e}, \phi) }\right)
\end{equation}
where $A$ is the cells' axon trajectory, $E$ is the set of electrodes, $\phi = \{\rho, \lambda, ...\}$ is a set of 12 patient-specific parameters,
and $d_s$ is the path length along the axon trajectory \cite{jansonius_mathematical_2009}from $\mathbf{a}$ to $\mathbf{x}$: 
\begin{equation} \label{eq:forward2}
    d_s(\mathbf{x}, \mathbf{a}) = \int_{\mathbf{a}}^\mathbf{x} \sqrt{A(\theta)^2 + \left(\frac{dA(\theta)}{d\theta}\right)^2}d\theta.
\end{equation}

This model ($f$) can be fit to individual patients; however, it is computationally slow and not differentiable.
For more details on these equations, see \citep{granley_computational_2021}. 
We therefore considered two approaches:
\begin{itemize}[topsep=0pt, leftmargin=15pt, parsep=0pt]
    \item \textbf{Differentiable Model:} We reformulated equations \ref{eq:forward1} and \ref{eq:forward2} into an equivalent set of parallelized matrix operations, implemented in Tensorflow \cite{tensorflow2015-whitepaper}. Significant efforts were put towards developing a model optimized for XLA engines on GPU, resulting in speedups of up to 5000x compared to the model as presented in \cite{granley_computational_2021}, enabling large-scale gradient descent. To enforce differentiability, we numerically approximated $d_s$ using $|A|=500$ axon segments per axon.
    \item \textbf{Surrogate Model:} We also implemented the surrogate approach from \cite{relic_deep_2022} as a baseline method, where $f$ is approximated with another \ac{DNN} $\hat{f}_\phi(\mathbf{s}; \theta_f)$ with weights $\theta_f$.
    To achieve this we generated 50,000 percepts using randomly selected stimuli passed through $f$ and fit a \ac{DNN} to produce identical images. As $f$ is highly dependent on patient-specific parameters $\phi$, we generated new data and fit a separate surrogate model for each $\phi$ in our experimental set.
    Specific implementation details of the surrogate are presented in Appendix \ref{app:surrogate}. Our implementation improves upon \cite{relic_deep_2022} by using the more advanced phosphene model described above, which accounts for effects of stimulus properties and allows for optimization of stimulus frequency in addition to amplitude.
\end{itemize}

\paragraph{Metrics} 
To measure perceptual similarity, we use a joint perceptual objective consisting of a VGG \citep{simonyan_very_2015-1} similarity term, a \acf{MAE} term, and a smoothness regularization term. The \ac{MAE} term is given by 
$L_\mathrm{MAE} = \frac{1}{|\mathbf{t}|}||\mathbf{t} - \hat{\mathbf{t}}||_1.$


The VGG term aims to capture higher-level differences between images \cite{de_ruyter_van_steveninck_end--end_2022, li_universal_2017}. The target image and reconstructed phosphene are input to VGG-19 pretrained on ImageNet \citep{deng_imagenet_2009}, and the MSE between the activations on a downstream convolutional layer is computed. 
Let $V_{l}$ be a function that extracts the activations of the $l$-th convolutional layer for a given image. The VGG loss is then defined as $L_\mathrm{VGG} = \frac{1}{|\mathbf{t}|}||V_{l}(\mathbf{t}) - V_{l}(\hat{\mathbf{t}})||_2^2$.

We also include a smoothing regularization term. This term imposes a loss on the second spatial derivative of the predicted image. A low second derivative implies that where the target image does change, it changes slowly. We found this encouraged smoother, more connected phosphenes. 
To approximate the second derivative, we convolve the image with a Laplacian filter \citep{paris_local_2015} of  size $k$, denoted by $Lap(\cdot, k)$, and compute the mean. The smoothness loss is given by:
\begin{equation} \label{eq:smooth}
    L_\mathrm{Smooth} = \frac{1}{|\hat{\mathbf{t}}|} \sum_i \Big(\frac{\partial^2}{dx^2} \hat{\mathbf{t}} \Big)_i
     =  \frac{1}{|\hat{\mathbf{t}}|} \sum_iLap(\hat{\mathbf{t}}, k)_i.
\end{equation}

Our final objective is the weighted sum of the three individual losses, 
given by Eq. \ref{eq:loss}, 
where $\alpha$ and $\beta$ are hyperparameters controlling the relative weighting of the three terms.
\begin{equation} \label{eq:loss}
    d = L_\mathrm{MAE} + \alpha L_\mathrm{Smooth} + \beta L_\mathrm{VGG}.
\end{equation}

We also implement a secondary metric to quantify phosphene recognizability, applicable only for the MNIST reconstruction task. We first pre-train a classifier network on the MNIST targets until it reaches 99\% test accuracy, and then fix the weights. The \ac{RA} is then defined as the ratio of the classifiers accuracy on the reconstructed images to its accuracy on the targets $RA = ACC / ACC(\mathbf{t})$. A perfect encoder would result in $RA=100\%$. A similar process was not possible for the COCO task due to the possibility of having multiple objects in each target image.
\paragraph{Training/Optimization} \label{sec:training}
We trained using Tensorflow 2.7 \cite{tensorflow2015-whitepaper} on a single NVIDIA RTX 3090 with 24GB memory. Stochastic gradient descent with Nesterov momentum was used to minimize the joint perceptual loss. We used a batch size of 16 due to memory limitations imposed by $f$. The amplitude, frequency predictions are scaled by 2, 20 respectively, while the pulse duration predictions were shifted by 1e-3 prior to being fed through the decoder. This encourages the initial predictions of the network to be in a reasonable range. The Laplacian filter size $k$ is set to 5. We choose $l$ to be first convolutional layer in the last block using cross validation (see Appendix \ref{app:hyperparameter}).
Similarly, we perform cross validation to find the best values for $\alpha$ and $\beta$. Instead of using one value, we found that incrementally increasing the weighting of the VGG loss ($\beta$) from 0 while simultaneously decreasing the initially high weight on the smoothing constraint ($\alpha$) was crucial for performance, especially when the range of allowed $\phi$ values was large (see Appendix \ref{app:hyperparameter}).

\paragraph{Datasets}
We first evaluated on handwritten digits from MNIST \cite{deng2012mnist}, enabling comparison to previous works \cite{relic_deep_2022}. Images preprocessing consisted of resizing the target images to the same shape as the output of $f$ (49x49). 
We also evaluate on more realistic images of common objects from the MS-COCO \citep{lin_microsoft_2015} dataset. We selected a subset of 25 of the MS-COCO object categories deemed more likely to be encountered by blind individuals (\textit{e.g.} people, household objects), and use only images that contain at least one instance of these objects. We further filter out images by various other criteria, such as being too cluttered or too dim. This process results in a total of approximately 47K training images and 12K test images. See Appendix \ref{app:coco_dataset} for a full description of the selection process. 

Natural images often contain too much detail to be encoded with prosthetic vision. While scene simplification strategies exist \citep{han_deep_2021}, we focus on the encoding algorithm, 
so we simply used the ground-truth segmentation masks to segment out the objects of interest. The images were then converted to grayscale, and resized to $49 \times 49$ pixels.

\begin{figure}
    \centering
    \includegraphics[width=\linewidth]{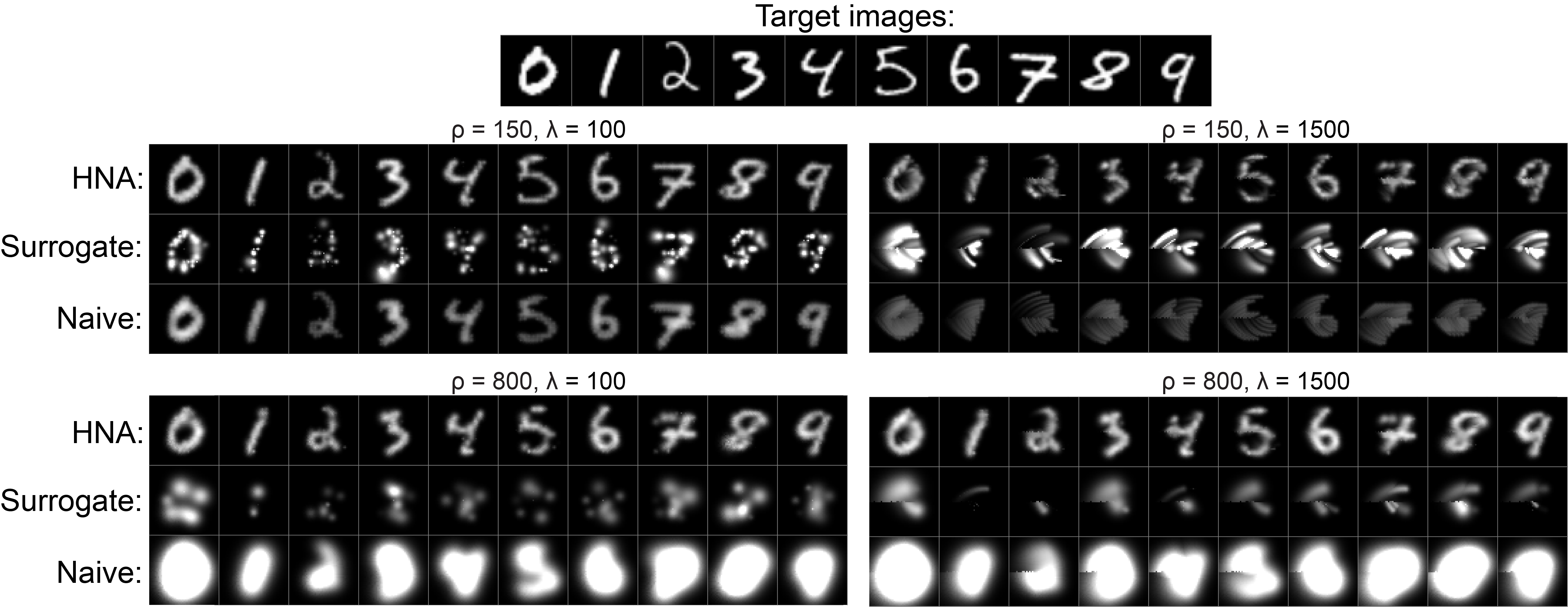}
    \caption{Reconstructed MNIST targets for \ac{HNA}, surrogate, and naive encoders across 4 specific simulated patients. Note that the brightness of the naive encoder is clipped for display}
    \label{fig:mnist-performance}
\end{figure}

\section{Results}
\subsection{MNIST} 
The phosphenes produced from the \ac{HNA}, surrogate, and naive encoders on the MNIST test set are shown in Fig.~\ref{fig:mnist-performance} and performance is summarized in Table~\ref{tab:mnist}. 
For each MNIST sample, the target image is input to the encoder, which predicts a stimulus. The stimulus is fed through the true forward model $f$, and the predicted phosphene is shown. Since the surrogate method must be retrained for each $\phi$, results are only shown for 4 simulated patients.
Our proposed approach outperformed the baselines across all metrics
(see Appendix \ref{app:stimuli} for a comparison of stimuli).

\begin{table}[t]
    \centering
    \vspace{6pt}
    \caption{MNIST performance}
    \label{tab:mnist}
    \resizebox{\columnwidth}{!}{%
    \begin{tabular}{@{}ccccccccccccc@{}}
        \toprule
        Encoding  & \multicolumn{3}{c}{$\rho$=150  $\lambda$=100} & \multicolumn{3}{c}{$\rho$=150  $\lambda$=1500} & \multicolumn{3}{c}{$\rho$=800  $\lambda$=100} & \multicolumn{3}{c}{$\rho$=800  $\lambda$=1500} \\ \midrule
                  & Joint Loss      & MAE & RA   & Joint Loss   & MAE    & RA     & Joint Loss     & MAE   & RA     & Joint Loss    & MAE     & RA         \\ \cmidrule(l){2-4} \cmidrule(l){5-7} \cmidrule(l){8-10} \cmidrule(l){11-13}
        Naive     & 1.161           & 0.1855 & 90.3 & 1.442        & 0.214  & 78.1   & 8.152          & 1.500 & 34.8   & 8.780         & 1.726      & 28.8    \\
        Surrogate & 2.509           & 0.1351 & 53.8 & 3.118        & 0.2431 &  30.7 & 1.692          & 0.2135  & 19.9 & 1.694         & 0.2237     &  18.1  \\
        HNA       & \textbf{0.559} & \textbf{0.064} & \textbf{98.1} & \textbf{1.029} & \textbf{0.1412} & \textbf{89.3} & \textbf{0.913} & \textbf{0.113} & \textbf{95.9} & \textbf{0.957}  & \textbf{0.126} & \textbf{94.8}\\ \bottomrule
    \end{tabular}
    }
\end{table}

\subsection{COCO} \label{sec:coco}

\begin{figure}[t]
    \centering
    \includegraphics[width=.9\linewidth]{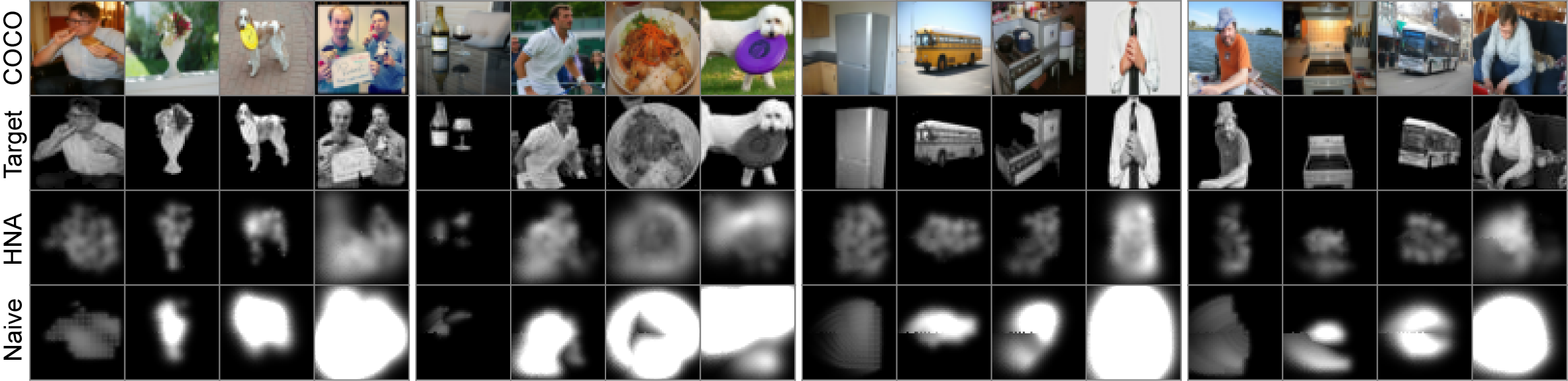}
    \caption{Original (\emph{top row}), segmented (\emph{second row}), and reconstructed targets for the COCO dataset, for both \ac{HNA} (\emph{third row}) and naive encoders (\emph{bottom row}). Left to right within each block of 4 images, $\rho$ takes values of 200, 400, 600, 800. Left to right across blocks, $\lambda$ takes values of 250, 750, 1250, 2000. Note that the brightness of the naive method is clipped for display.}
    \label{fig:coco}
\end{figure}

The phosphenes produced by \ac{HNA} and the naive encoder for the segmented COCO dataset are shown in Fig.~\ref{fig:coco}. We omit the surrogate results due to its poor perceptual performance on MNIST. Averaged across all $\phi$, \ac{HNA} had a joint loss of 0.713 on the test set and MAE of 0.1408, while the naive encoder had a joint loss of 1.873 and \ac{MAE} of 0.2830.

\begin{figure}[t]
    \centering
    \includegraphics[width=0.95\linewidth]{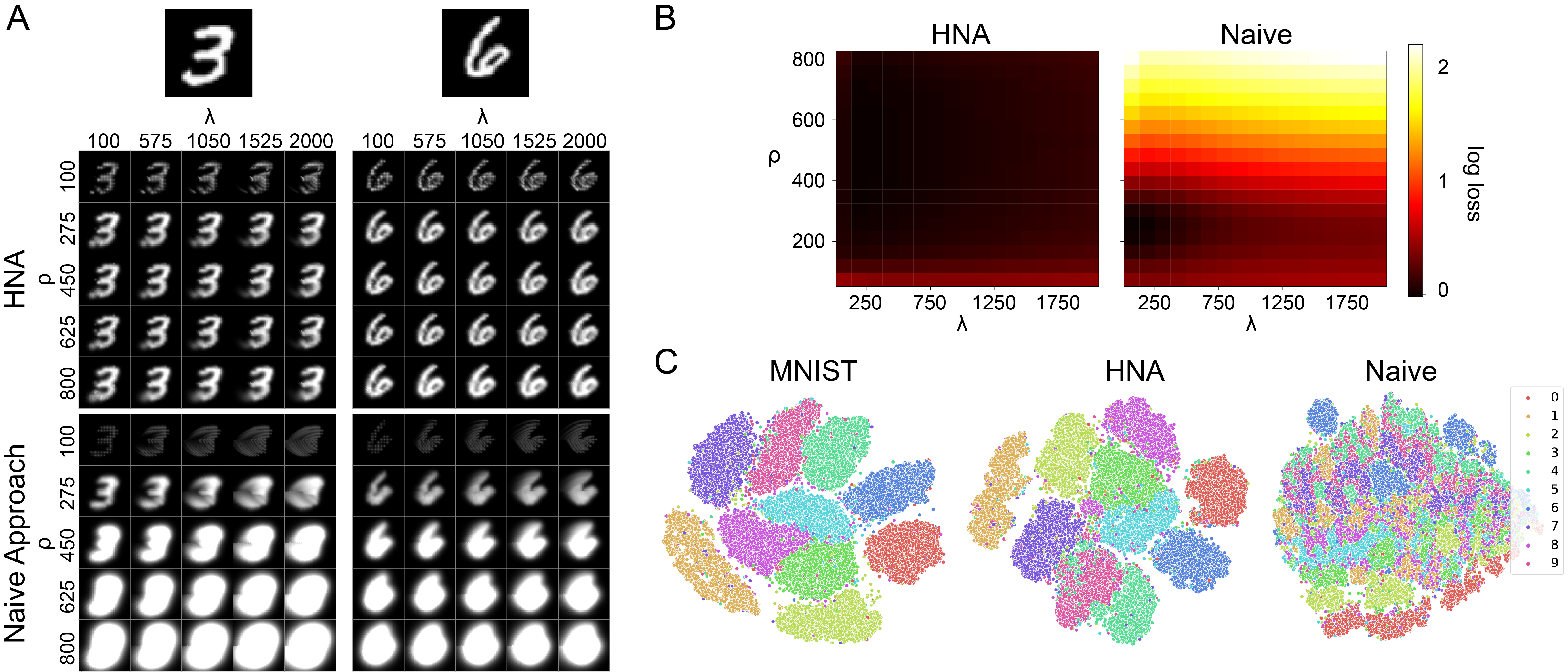}
    \caption{Encoder performance across simulated patients (varying $\rho$ and $\lambda$) on the MNIST dataset. \textbf{\textit{A}}: Target, \ac{HNA} encoder, and naive encoder phosphenes for two example digits. \textbf{\textit{B}}: Heatmaps showing the log joint loss across $\rho$ and $\lambda$ for \ac{HNA} and naive encoders. \textbf{\textit{C}}: T-SNE clusterings on original MNIST targets, \ac{HNA} reconstructed phosphenes, and naive reconstructed phosphenes.}
    \label{fig:mnist-rho-lam}
\end{figure}

\subsection{Modeling Patient-to-Patient Variations} \label{sec:patient-to-patient}
MNIST encoder performance across simulated patients ($\phi$) is shown in Fig.~\ref{fig:mnist-rho-lam}.
Since the surrogate encoder has to be retrained for each patient, comparison is infeasible. 
To visualize the effects of changing $\rho$ and $\lambda$ on the produced phosphenes, Fig.~\ref{fig:mnist-rho-lam}A shows the result of encoding two example MNIST digits, both using the naive method and our encoder. As $\lambda$ increases, the naive phosphenes appear increasingly elongated, and as $\rho$ increases, the phosphenes become increasingly large and blurry. The phosphenes from \ac{HNA} are slightly too dim and disconnected at low $\rho$, but are relatively stable across other values of $\rho$ and $\lambda$. 

To compare performance across the entire dataset, we computed the average test set loss across the same range of $\rho$ and $\lambda$
(Fig.~\ref{fig:mnist-rho-lam}B).
%
The encoder performs well across a wide range of simulated patients, with larger loss only at low $\rho$. The naive method performs well only on a limited set of $\phi$, with small $\lambda$ and $\rho \approx 200$. The naive loss was higher than the learned encoder at every simulated point. Random sampling of $\rho$ and $\lambda$ for each image results in a joint loss of 0.921, \ac{MAE} of 0.120, and \ac{RA} of 94.0\% for \ac{HNA}, while the naive encoder results in a joint loss of 3.17, \ac{MAE} of 0.596, and \ac{RA} of 63.6\%. 
The same analysis yielded similar results on COCO (Appendix \ref{app:coco_patient}). An analysis across other parameters is presented in Appendix \ref{app:other_modelparams}.

In order for prosthetic vision to be useful, different instances of the same objects would ideally produce similar phosphenes, allowing for consistent perception. To evaluate whether our model achieves this, we cluster the target images and resulting phosphenes using t-SNE \citep{van2008visualizing} shown in Fig.~\ref{fig:mnist-rho-lam}C. The ground truth images form clusters corresponding to the digits 0-9. The phosphenes from our encoder roughly form similar, slightly less separated groupings, whereas the naive phosphenes do not.
To ensure that this was not the result of bad t-SNE hyperparameters, we repeated the clustering across different perplexities and learning rates, obtaining similar or worse results.


\subsection{Joint Perceptual Error Ablation Study} 
\begin{figure}[t]
    \centering
    \includegraphics[width=0.9\linewidth]{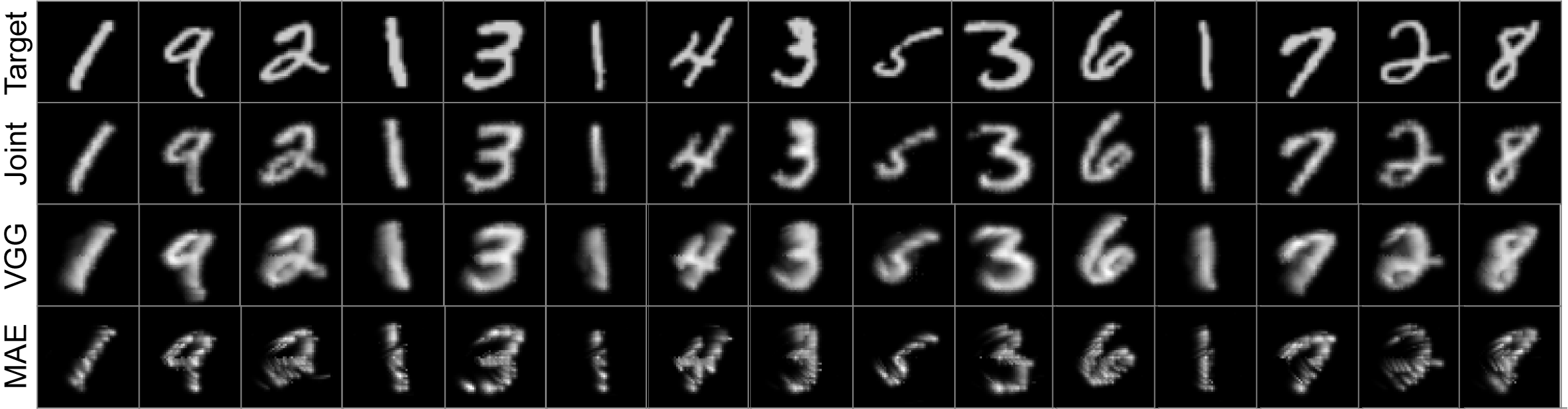}
    \caption{MNIST images for \ac{HNA} encoders trained using the joint, VGG-only, and MAE-only loss.}
    \label{fig:ablation}
\end{figure}

To show that the joint perceptual metric performs better than any of its individual components, we train models using just the VGG loss and just MAE loss. Shown are values for $\rho$=150 and $\lambda$=600. As mentioned previously, encoders trained using just VGG loss fail to converge, thus we pretrain the VGG encoder using \ac{MAE} and smoothing loss, then transition to using only VGG. We do not consider ablating the smoothing term (Eq. \ref{eq:smooth}) because it is simply a regularization term.
Fig.~\ref{fig:ablation} shows the phosphenes produced by \ac{HNA} trained on the joint, VGG-only, and MAE-only loss.

The VGG encoder had a test VGG loss of 4\% lower than the joint model, but its produced phosphenes are oversmoothed and blurry. The \ac{MAE} encoder had a final test \ac{MAE} of 9\% lower than the joint model, but its produced phosphenes are disconnected and low-quality. The joint model had a \ac{RA} of 99.0\%, the VGG encoder had a \ac{RA} of 95.9\%, and the joint model had a \ac{RA} of 77.6\%

\section{Discussion} \label{sec:discussion}


\paragraph{Visual Prostheses}
We found that \ac{HNA} is able to produce high-fidelity stimuli from the MNIST and COCO datasets that outperform conventional encoding strategies across all tested conditions. Importantly, \ac{HNA} produces phosphenes that are consistent across representations of the same object (Fig.~\ref{fig:mnist-rho-lam}C), which is critical to allowing prosthesis users to learn to associate certain visual patterns with specific objects.
On the MNIST task, \ac{HNA} produced high quality reconstructions, nearly matching the targets (Figure \ref{fig:mnist-performance}). On the harder COCO task, \ac{HNA} significantly outperformed the naive encoder, but was still unable to capture all of the detail in the images.
In Appendix \ref{app:high-res}, we demonstrate that this is largely due to the implant's limited spatial resolution and not a fundamental limitation of \ac{HNA}.

Another advantage of the \ac{HNA} is that it can be trained to predict stimuli across a wide range of patient-specific parameter values $\phi$, whereas the conventional naive encoder works well only for small values of $\rho$ and $\lambda$.
This may be one reason why the naive encoding strategy has been shown to lead to substantial individual differences in visual outcomes \cite{stingl_interim_2017,peli_testing_2020}.
Our results suggest that stimuli produced with \ac{HNA} may be able to reduce at least some amount of this patient-to-patient variability.

Furthermore, \ac{HNA} also proved superior to a surrogate forward model.
The latter offer an alternative when the forward model is computationally expensive or not differentiable.
Understandably, any inaccuracies in the surrogate model will propagate to the learned encoder during training. However, we observed 
that even for well trained surrogates, the encoder may still learn to exploit the inexact surrogate instead of learning to invert the true model (see Appendix~\ref{app:surrogate}). It is possible that this exploitation could be mitigated to some extent by adversarially-robust training techniques \cite{tramer2017ensemble}. We suspect that the surrogate method's inferior performance here compared to \cite{relic_deep_2022} can be explained by our larger stimulus search space. Thus, we cannot currently suggest \ac{HNA} for surrogate forward models, unless the forward model is sufficiently simple or has a small stimulus space.



\paragraph{Deployment} \ac{HNA} encoders must be lightweight enough to be deployed in resource-limited neuroprosthetic environments. Our encoder's single image inference time was 1.2ms on GPU and 4ms on CPU. Future work could reduce these numbers through network pruning, mixed precision, and architecture search. Low-power Edge AI accelerators (\emph{e.g.}, Intel’s Neural Compute Stick) and dedicated neuromorphic hardware (\emph{e.g.}, BrainChip’s Akida SoC) may provide another solution.


\paragraph{Broader Impacts}

While our work is presented in the context of visual prostheses, the \ac{HNA} framework may apply to any sensory neuroprosthesis where stimulus selection can be informed by a numeric or symbolic forward model.
For example, \ac{HNA} could be used in cochlear implants \cite{wilson_better_1991} to choose stimuli that result in a desired sound, and in spinal cord implants \citep{capogrosso_brainspine_2016} to find the best way to relay neural signals through a damaged section of the spinal cord.
Conveniently, the forward models required by \ac{HNA} have already been developed for a range of applications \cite{dorman2005acoustic, svirsky2013validation, dorman2003simulations, dorman1997speech, cooper2008music, loizou2000speech, saal2017simulating, okorokova2018biomimetic, weber2012interfacing, kim2010predicting, mileusnic2006mathematical}.
However, \ac{HNA} might not apply to all neural interfaces, such as systems without a clear neural or perceptual target (\textit{e.g.}, deep brain stimulation for the treatment of Parkinson's \citep{benabid_deep_2003}) or closed-loop systems \citep{bozorgzadeh_neurochemostat_2016, chapman_multifunctional_2018}.

\paragraph{Limitations}
Despite \ac{HNA}'s potential, the current implementation has a number of limitations.
First, as presented the HNA encoder only applies to static targets. Hence dynamic targets must be split into individual frames and encoded separately. However, one approach might be to encode entire stimulus sequences (instead of frames) that are optimized to reconstruct the dynamic target sequence.

Second, HNA works best if there is an accurate forward model mapping from stimulus space to perception. However, Appendix~\ref{app:mis-specified} shows that HNA may still give benefits over a naive encoding even when patient-specific parameters are unknown or mis-specified. In general, if a prosthesis elicits similar results across patients, then a non-patient-specific model would suffice.

Third, the current works deals only with simulated patients.
 The use of a \ac{DNN} for stimulus encoding  in real patients may raise safety concerns. 
 Since we cannot examine the process by which stimuli are chosen, it is possible that \ac{HNA} might produce harmful stimuli that could lead to serious adverse events (\textit{e.g.}, seizures).
 However, this concern is mitigated by the fact that most neuroprostheses are equipped with firmware responsible for ensuring stimuli stay within FDA-approved safety limits. 

\section{Conclusion}

In summary, this paper proposes a hybrid autoencoder structure as a general framework for stimulus optimization in sensory neuroprostheses and, as a proof of concept, demonstrates its utility on the prominent example of visual neuroprostheses, drastically outperforming conventional encoding strategies.
This constitutes an important step towards the long-standing challenge of restoring high-quality vision to people living with incurable blindness and may prove a promising solution for a variety of neuroprosthetic technologies.

\section{Acknowledgements}
This work was supported by the National Institutes of Health (NIH R00 EY-029329 to MB).


\newpage
{
\small

\bibliography{refs.bib}

\begin{thebibliography}{10}

\bibitem{cinel_neurotechnologies_2019}
Caterina Cinel, Davide Valeriani, and Riccardo Poli.
\newblock Neurotechnologies for {Human} {Cognitive} {Augmentation}: {Current}
  {State} of the {Art} and {Future} {Prospects}.
\newblock {\em Frontiers in Human Neuroscience}, 13:13, January 2019.

\bibitem{fernandez_development_2018}
Eduardo Fernandez.
\newblock Development of visual {Neuroprostheses}: trends and challenges.
\newblock {\em Bioelectronic Medicine}, 4(1):12, August 2018.

\bibitem{wilson_better_1991}
Blake~S. Wilson, Charles~C. Finley, Dewey~T. Lawson, Robert~D. Wolford,
  Donald~K. Eddington, and William~M. Rabinowitz.
\newblock Better speech recognition with cochlear implants.
\newblock {\em Nature}, 352(6332):236--238, July 1991.
\newblock Number: 6332 Publisher: Nature Publishing Group.

\bibitem{fernandez_visual_2021}
Eduardo Fernández, Arantxa Alfaro, Cristina Soto-Sánchez, Pablo
  Gonzalez-Lopez, Antonio~M. Lozano, Sebastian Peña, Maria~Dolores Grima,
  Alfonso Rodil, Bernardeta Gómez, Xing Chen, Pieter~R. Roelfsema, John~D.
  Rolston, Tyler~S. Davis, and Richard~A. Normann.
\newblock Visual percepts evoked with an intracortical 96-channel
  microelectrode array inserted in human occipital cortex.
\newblock {\em Journal of Clinical Investigation}, 131(23):e151331, December
  2021.

\bibitem{luo_argus_2016}
Yvonne Hsu-Lin Luo and Lyndon da~Cruz.
\newblock The {Argus}® {II} {Retinal} {Prosthesis} {System}.
\newblock {\em Progress in Retinal and Eye Research}, 50:89--107, January 2016.

\bibitem{tan_neural_2014}
Daniel~W. Tan, Matthew~A. Schiefer, Michael~W. Keith, James~Robert Anderson,
  Joyce Tyler, and Dustin~J. Tyler.
\newblock A neural interface provides long-term stable natural touch
  perception.
\newblock {\em Science Translational Medicine}, 6(257):257ra138--257ra138,
  October 2014.
\newblock Publisher: American Association for the Advancement of Science.

\bibitem{erickson-davis_what_2021}
Cordelia Erickson-Davis and Helma Korzybska.
\newblock What do blind people “see” with retinal prostheses?
  {Observations} and qualitative reports of epiretinal implant users.
\newblock {\em PLOS ONE}, 16(2):e0229189, February 2021.
\newblock Publisher: Public Library of Science.

\bibitem{murray_embodiment_2008}
Craig~D. Murray.
\newblock Embodiment and {Prosthetics}.
\newblock In Pamela Gallagher, Deirdre Desmond, and Malcolm MacLachlan,
  editors, {\em Psychoprosthetics}, pages 119--129. Springer, London, 2008.

\bibitem{brunton_data-driven_2019}
Bingni~W. Brunton and Michael Beyeler.
\newblock Data-driven models in human neuroscience and neuroengineering.
\newblock {\em Current Opinion in Neurobiology}, 58:21--29, October 2019.

\bibitem{relic_deep_2022}
Lucas Relic, Bowen Zhang, Yi-Lin Tuan, and Michael Beyeler.
\newblock Deep {Learning}--{Based} {Perceptual} {Stimulus} {Encoder} for
  {Bionic} {Vision}.
\newblock In {\em Augmented {Humans} 2022}, {AHs} 2022, pages 323--325, New
  York, NY, USA, March 2022. Association for Computing Machinery.

\bibitem{montes_de_oca_zapiain_accelerating_2021}
David Montes~de Oca~Zapiain, James~A. Stewart, and Rémi Dingreville.
\newblock Accelerating phase-field-based microstructure evolution predictions
  via surrogate models trained by machine learning methods.
\newblock {\em npj Computational Materials}, 7(1):1--11, January 2021.
\newblock Number: 1 Publisher: Nature Publishing Group.

\bibitem{nabian_deep_2019}
Mohammad~Amin Nabian and Hadi Meidani.
\newblock A {Deep} {Neural} {Network} {Surrogate} for {High}-{Dimensional}
  {Random} {Partial} {Differential} {Equations}.
\newblock {\em Probabilistic Engineering Mechanics}, 57:14--25, July 2019.
\newblock arXiv:1806.02957 [physics, stat].

\bibitem{nikolopoulos_non-intrusive_2022}
Stefanos Nikolopoulos, Ioannis Kalogeris, and Vissarion Papadopoulos.
\newblock Non-intrusive surrogate modeling for parametrized time-dependent
  partial differential equations using convolutional autoencoders.
\newblock {\em Engineering Applications of Artificial Intelligence},
  109:104652, March 2022.

\bibitem{tabot_restoring_2013}
Gregg~A. Tabot, John~F. Dammann, Joshua~A. Berg, Francesco~V. Tenore,
  Jessica~L. Boback, R.~Jacob Vogelstein, and Sliman~J. Bensmaia.
\newblock Restoring the sense of touch with a prosthetic hand through a brain
  interface.
\newblock {\em Proceedings of the National Academy of Sciences},
  110(45):18279--18284, November 2013.

\bibitem{capogrosso_brainspine_2016}
Marco Capogrosso, Tomislav Milekovic, David Borton, Fabien Wagner,
  Eduardo~Martin Moraud, Jean-Baptiste Mignardot, Nicolas Buse, Jerome Gandar,
  Quentin Barraud, David Xing, Elodie Rey, Simone Duis, Yang Jianzhong, Wai
  Kin~D. Ko, Qin Li, Peter Detemple, Tim Denison, Silvestro Micera, Erwan
  Bezard, Jocelyne Bloch, and Grégoire Courtine.
\newblock A brain–spine interface alleviating gait deficits after spinal cord
  injury in primates.
\newblock {\em Nature}, 539(7628):284--288, November 2016.

\bibitem{chapman_multifunctional_2018}
Christopher A.~R. Chapman, Noah Goshi, and Erkin Seker.
\newblock Multifunctional {Neural} {Interfaces} for {Closed}-{Loop} {Control}
  of {Neural} {Activity}.
\newblock {\em Advanced Functional Materials}, 28(12):1703523, 2018.
\newblock \_eprint:
  https://onlinelibrary.wiley.com/doi/pdf/10.1002/adfm.201703523.

\bibitem{wagner_targeted_2018}
Fabien~B. Wagner, Jean-Baptiste Mignardot, Camille~G. Le~Goff-Mignardot, Robin
  Demesmaeker, Salif Komi, Marco Capogrosso, Andreas Rowald, Ismael Seáñez,
  Miroslav Caban, Elvira Pirondini, Molywan Vat, Laura~A. McCracken, Roman
  Heimgartner, Isabelle Fodor, Anne Watrin, Perrine Seguin, Edoardo Paoles,
  Katrien Van Den~Keybus, Grégoire Eberle, Brigitte Schurch, Etienne Pralong,
  Fabio Becce, John Prior, Nicholas Buse, Rik Buschman, Esra Neufeld, Niels
  Kuster, Stefano Carda, Joachim von Zitzewitz, Vincent Delattre, Tim Denison,
  Hendrik Lambert, Karen Minassian, Jocelyne Bloch, and Grégoire Courtine.
\newblock Targeted neurotechnology restores walking in humans with spinal cord
  injury.
\newblock {\em Nature}, 563(7729):65--71, November 2018.

\bibitem{ayton_update_2020}
Lauren~N. Ayton, Nick Barnes, Gislin Dagnelie, Takashi Fujikado, Georges Goetz,
  Ralf Hornig, Bryan~W. Jones, Mahiul M.~K. Muqit, Daniel~L. Rathbun, Katarina
  Stingl, James~D. Weiland, and Matthew~A. Petoe.
\newblock An update on retinal prostheses.
\newblock {\em Clinical Neurophysiology}, 131(6):1383--1398, June 2020.

\bibitem{stingl_interim_2017}
Katarina Stingl, Ruth Schippert, Karl~U. Bartz-Schmidt, Dorothea Besch,
  Charles~L. Cottriall, Thomas~L. Edwards, Florian Gekeler, Udo Greppmaier,
  Katja Kiel, Assen Koitschev, Laura Kühlewein, Robert~E. MacLaren, James~D.
  Ramsden, Johann Roider, Albrecht Rothermel, Helmut Sachs, Greta~S. Schröder,
  Jan Tode, Nicole Troelenberg, and Eberhart Zrenner.
\newblock Interim {Results} of a {Multicenter} {Trial} with the {New}
  {Electronic} {Subretinal} {Implant} {Alpha} {AMS} in 15 {Patients} {Blind}
  from {Inherited} {Retinal} {Degenerations}.
\newblock {\em Frontiers in Neuroscience}, 11, 2017.
\newblock Publisher: Frontiers.

\bibitem{karapanos_functional_2021}
Lewis Karapanos, Carla~J. Abbott, Lauren~N. Ayton, Maria Kolic, Myra~B.
  McGuinness, Elizabeth~K. Baglin, Samuel~A. Titchener, Jessica Kvansakul, Dean
  Johnson, William~G. Kentler, Nick Barnes, David A.~X. Nayagam, Penelope~J.
  Allen, and Matthew~A. Petoe.
\newblock Functional {Vision} in the {Real}-{World} {Environment} {With} a
  {Second}-{Generation} (44-{Channel}) {Suprachoroidal} {Retinal} {Prosthesis}.
\newblock {\em Translational Vision Science \& Technology}, 10(10):7--7, August
  2021.
\newblock Publisher: The Association for Research in Vision and Ophthalmology.

\bibitem{dobelle_artificial_2000}
Wm~H. Dobelle.
\newblock Artificial {Vision} for the {Blind} by {Connecting} a {Television}
  {Camera} to the {Visual} {Cortex}.
\newblock {\em ASAIO Journal}, 46(1):3--9, February 2000.

\bibitem{beyeler_model_2019}
Michael Beyeler, Devyani Nanduri, James~D. Weiland, Ariel Rokem, Geoffrey~M.
  Boynton, and Ione Fine.
\newblock A model of ganglion axon pathways accounts for percepts elicited by
  retinal implants.
\newblock {\em Scientific Reports}, 9(1):1--16, June 2019.

\bibitem{rizzo_perceptual_2003}
J.~F. Rizzo, J.~Wyatt, J.~Loewenstein, S.~Kelly, and D.~Shire.
\newblock Perceptual efficacy of electrical stimulation of human retina with a
  microelectrode array during short-term surgical trials.
\newblock {\em Invest Ophthalmol Vis Sci}, 44(12):5362--9, December 2003.

\bibitem{granley_computational_2021}
Jacob Granley and Michael Beyeler.
\newblock A {Computational} {Model} of {Phosphene} {Appearance} for
  {Epiretinal} {Prostheses}.
\newblock In {\em 2021 43rd {Annual} {International} {Conference} of the {IEEE}
  {Engineering} in {Medicine} {Biology} {Society} ({EMBC})}, pages 4477--4481,
  November 2021.
\newblock ISSN: 2694-0604.

\bibitem{luo_long-term_2016}
Yvonne H-L. Luo, Joe~Jiangjian Zhong, Monica Clemo, and Lyndon da~Cruz.
\newblock Long-term {Repeatability} and {Reproducibility} of {Phosphene}
  {Characteristics} in {Chronically} {Implanted} {Argus} {II} {Retinal}
  {Prosthesis} {Subjects}.
\newblock {\em American Journal of Ophthalmology}, 170:100--109, October 2016.

\bibitem{beyeler_learning_2017}
M.~Beyeler, A.~Rokem, G.~M. Boynton, and I.~Fine.
\newblock Learning to see again: biological constraints on cortical plasticity
  and the implications for sight restoration technologies.
\newblock {\em J Neural Eng}, 14(5):051003, June 2017.

\bibitem{chen_simulating_2009}
S.~C. Chen, G.~J. Suaning, J.~W. Morley, and N.~H. Lovell.
\newblock Simulating prosthetic vision: {I}. {Visual} models of phosphenes.
\newblock {\em Vision Research}, 49(12):1493--506, June 2009.

\bibitem{tong2020stimulation}
Wei Tong, Hamish Meffin, David~J Garrett, and Michael~R Ibbotson.
\newblock Stimulation strategies for improving the resolution of retinal
  prostheses.
\newblock {\em Frontiers in neuroscience}, 14:262, 2020.

\bibitem{shah_optimization_2019}
Nishal~P. Shah, Sasidhar Madugula, Lauren Grosberg, Gonzalo Mena, Pulkit
  Tandon, Pawel Hottowy, Alexander Sher, Alan Litke, Subhasish Mitra, and E.J.
  Chichilnisky.
\newblock Optimization of {Electrical} {Stimulation} for a {High}-{Fidelity}
  {Artificial} {Retina}.
\newblock In {\em 2019 9th {International} {IEEE}/{EMBS} {Conference} on
  {Neural} {Engineering} ({NER})}, pages 714--718, March 2019.
\newblock ISSN: 1948-3554.

\bibitem{ghaffari_closed-loop_2021}
Dorsa~Haji Ghaffari, Yao-Chuan Chang, Ehsan Mirzakhalili, and James~D. Weiland.
\newblock Closed-loop {Optimization} of {Retinal} {Ganglion} {Cell} {Responses}
  to {Epiretinal} {Stimulation}: {A} {Computational} {Study}.
\newblock In {\em 2021 10th {International} {IEEE}/{EMBS} {Conference} on
  {Neural} {Engineering} ({NER})}, pages 597--600, May 2021.
\newblock ISSN: 1948-3554.

\bibitem{fauvel_human---loop_2021}
Tristan Fauvel and Matthew Chalk.
\newblock Human-in-the-loop optimization of visual prosthetic stimulation.
\newblock preprint, Neuroscience, November 2021.

\bibitem{spencer_global_2019}
Martin~J. Spencer, Tatiana Kameneva, David~B. Grayden, Hamish Meffin, and
  Anthony~N. Burkitt.
\newblock Global activity shaping strategies for a retinal implant.
\newblock {\em Journal of Neural Engineering}, 16(2):026008, January 2019.
\newblock Publisher: IOP Publishing.

\bibitem{de_ruyter_van_steveninck_end--end_2022}
Jaap de~Ruyter~van Steveninck, Umut Güçlü, Richard van Wezel, and Marcel van
  Gerven.
\newblock End-to-end optimization of prosthetic vision.
\newblock {\em Journal of Vision}, 22(2):20, February 2022.

\bibitem{marc_remodeling_2003}
Robert~E Marc, Bryan~W Jones, Carl~B Watt, and Enrica Strettoi.
\newblock Neural remodeling in retinal degeneration.
\newblock {\em Progress in Retinal and Eye Research}, 22(5):607--655, 2003.

\bibitem{wang_mean_2009}
Zhou Wang and Alan~C. Bovik.
\newblock Mean squared error: {Love} it or leave it? {A} new look at {Signal}
  {Fidelity} {Measures}.
\newblock {\em IEEE Signal Processing Magazine}, 26(1):98--117, January 2009.
\newblock Conference Name: IEEE Signal Processing Magazine.

\bibitem{zhai_perceptual_2020}
Guangtao Zhai and Xiongkuo Min.
\newblock Perceptual image quality assessment: a survey.
\newblock {\em Science China Information Sciences}, 63(11):211301, November
  2020.

\bibitem{ioffe_batch_2015}
Sergey Ioffe and Christian Szegedy.
\newblock Batch {Normalization}: {Accelerating} {Deep} {Network} {Training} by
  {Reducing} {Internal} {Covariate} {Shift}.
\newblock Technical Report arXiv:1502.03167, arXiv, March 2015.
\newblock arXiv:1502.03167 [cs] type: article.

\bibitem{dorman2005acoustic}
Michael~F Dorman, Anthony~J Spahr, Philipos~C Loizou, Cindy~J Dana, and
  Jennifer~S Schmidt.
\newblock Acoustic simulations of combined electric and acoustic hearing (eas).
\newblock {\em Ear and Hearing}, 26(4):371--380, 2005.

\bibitem{svirsky2013validation}
Mario~A Svirsky, Nai Ding, Elad Sagi, Chin-Tuan Tan, Matthew Fitzgerald,
  E~Katelyn Glassman, Keena Seward, and Arlene~C Neuman.
\newblock Validation of acoustic models of auditory neural prostheses.
\newblock In {\em 2013 IEEE International Conference on Acoustics, Speech and
  Signal Processing}, pages 8629--8633. IEEE, 2013.

\bibitem{dorman2003simulations}
MF~Dorman, PC~Loizou, A~Spahr, and CJ~Dana.
\newblock Simulations of combined acoustic/electric hearing.
\newblock In {\em Proceedings of the 25th Annual International Conference of
  the IEEE Engineering in Medicine and Biology Society (IEEE Cat. No.
  03CH37439)}, volume~3, pages 1999--2001. IEEE, 2003.

\bibitem{dorman1997speech}
Michael~F Dorman, Philipos~C Loizou, and Dawne Rainey.
\newblock Speech intelligibility as a function of the number of channels of
  stimulation for signal processors using sine-wave and noise-band outputs.
\newblock {\em The Journal of the Acoustical Society of America},
  102(4):2403--2411, 1997.

\bibitem{cooper2008music}
William~B Cooper, Emily Tobey, and Philipos~C Loizou.
\newblock Music perception by cochlear implant and normal hearing listeners as
  measured by the montreal battery for evaluation of amusia.
\newblock {\em Ear and hearing}, 29(4):618, 2008.

\bibitem{loizou2000speech}
Philipos~C Loizou, Michael Dorman, Oguz Poroy, and Tony Spahr.
\newblock Speech recognition by normal-hearing and cochlear implant listeners
  as a function of intensity resolution.
\newblock {\em The Journal of the Acoustical Society of America},
  108(5):2377--2387, 2000.

\bibitem{saal2017simulating}
Hannes~P Saal, Benoit~P Delhaye, Brandon~C Rayhaun, and Sliman~J Bensmaia.
\newblock Simulating tactile signals from the whole hand with millisecond
  precision.
\newblock {\em Proceedings of the National Academy of Sciences},
  114(28):E5693--E5702, 2017.

\bibitem{okorokova2018biomimetic}
Elizaveta~V Okorokova, Qinpu He, and Sliman~J Bensmaia.
\newblock Biomimetic encoding model for restoring touch in bionic hands through
  a nerve interface.
\newblock {\em Journal of neural engineering}, 15(6):066033, 2018.

\bibitem{weber2012interfacing}
Douglas~J Weber, Rebecca Friesen, and Lee~E Miller.
\newblock Interfacing the somatosensory system to restore touch and
  proprioception: essential considerations.
\newblock {\em Journal of motor behavior}, 44(6):403--418, 2012.

\bibitem{kim2010predicting}
Sung~Soo Kim, Arun~P Sripati, and Sliman~J Bensmaia.
\newblock Predicting the timing of spikes evoked by tactile stimulation of the
  hand.
\newblock {\em Journal of neurophysiology}, 104(3):1484--1496, 2010.

\bibitem{mileusnic2006mathematical}
Milana~P Mileusnic, Ian~E Brown, Ning Lan, and Gerald~E Loeb.
\newblock Mathematical models of proprioceptors. i. control and transduction in
  the muscle spindle.
\newblock {\em Journal of neurophysiology}, 96(4):1772--1788, 2006.

\bibitem{jansonius_mathematical_2009}
N.~M. Jansonius, J.~Nevalainen, B.~Selig, L.~M. Zangwill, P.~A. Sample, W.~M.
  Budde, J.~B. Jonas, W.~A. Lagrèze, P.~J. Airaksinen, R.~Vonthein, L.~A.
  Levin, J.~Paetzold, and U.~Schiefer.
\newblock A mathematical description of nerve fiber bundle trajectories and
  their variability in the human retina.
\newblock {\em Vision Research}, 49(17):2157--2163, August 2009.

\bibitem{tensorflow2015-whitepaper}
Mart\'{i}n Abadi, Ashish Agarwal, Paul Barham, Eugene Brevdo, Zhifeng Chen,
  Craig Citro, Greg~S. Corrado, Andy Davis, Jeffrey Dean, Matthieu Devin,
  Sanjay Ghemawat, Ian Goodfellow, Andrew Harp, Geoffrey Irving, Michael Isard,
  Yangqing Jia, Rafal Jozefowicz, Lukasz Kaiser, Manjunath Kudlur, Josh
  Levenberg, Dandelion Man\'{e}, Rajat Monga, Sherry Moore, Derek Murray, Chris
  Olah, Mike Schuster, Jonathon Shlens, Benoit Steiner, Ilya Sutskever, Kunal
  Talwar, Paul Tucker, Vincent Vanhoucke, Vijay Vasudevan, Fernanda Vi\'{e}gas,
  Oriol Vinyals, Pete Warden, Martin Wattenberg, Martin Wicke, Yuan Yu, and
  Xiaoqiang Zheng.
\newblock {TensorFlow}: Large-scale machine learning on heterogeneous systems,
  2015.
\newblock Software available from tensorflow.org.

\bibitem{simonyan_very_2015-1}
Karen Simonyan and Andrew Zisserman.
\newblock Very {Deep} {Convolutional} {Networks} for {Large}-{Scale} {Image}
  {Recognition}.
\newblock Technical Report arXiv:1409.1556, arXiv, April 2015.
\newblock arXiv:1409.1556 [cs] type: article.

\bibitem{li_universal_2017}
Yijun Li, Chen Fang, Jimei Yang, Zhaowen Wang, Xin Lu, and Ming-Hsuan Yang.
\newblock Universal {Style} {Transfer} via {Feature} {Transforms}.
\newblock In {\em Advances in {Neural} {Information} {Processing} {Systems}},
  volume~30. Curran Associates, Inc., 2017.

\bibitem{deng_imagenet_2009}
Jia Deng, Wei Dong, Richard Socher, Li-Jia Li, Kai Li, and Li~Fei-Fei.
\newblock {ImageNet}: {A} large-scale hierarchical image database.
\newblock In {\em 2009 {IEEE} {Conference} on {Computer} {Vision} and {Pattern}
  {Recognition}}, pages 248--255, June 2009.
\newblock ISSN: 1063-6919.

\bibitem{paris_local_2015}
Sylvain Paris, Samuel~W Hasinoff, and Jan Kautz.
\newblock Local {Laplacian} {Filters}: {Edge}-aware {Image} {Processing} with a
  {Laplacian} {Pyramid}.
\newblock {\em Communications of the ACM}, 58:11, 2015.

\bibitem{deng2012mnist}
Li~Deng.
\newblock The mnist database of handwritten digit images for machine learning
  research.
\newblock {\em IEEE Signal Processing Magazine}, 29(6):141--142, 2012.

\bibitem{lin_microsoft_2015}
Tsung-Yi Lin, Michael Maire, Serge Belongie, Lubomir Bourdev, Ross Girshick,
  James Hays, Pietro Perona, Deva Ramanan, C.~Lawrence Zitnick, and Piotr
  Dollár.
\newblock Microsoft {COCO}: {Common} {Objects} in {Context}.
\newblock Technical Report arXiv:1405.0312, arXiv, February 2015.
\newblock arXiv:1405.0312 [cs] type: article.

\bibitem{han_deep_2021}
Nicole Han, Sudhanshu Srivastava, Aiwen Xu, Devi Klein, and Michael Beyeler.
\newblock Deep {Learning}--{Based} {Scene} {Simplification} for {Bionic}
  {Vision}.
\newblock In {\em Augmented {Humans} {Conference} 2021}, {AHs}'21, pages
  45--54, New York, NY, USA, February 2021. Association for Computing
  Machinery.

\bibitem{van2008visualizing}
Laurens Van~der Maaten and Geoffrey Hinton.
\newblock Visualizing data using t-sne.
\newblock {\em Journal of machine learning research}, 9(11), 2008.

\bibitem{peli_testing_2020}
Eli Peli.
\newblock Testing {Vision} {Is} {Not} {Testing} {For} {Vision}.
\newblock {\em Translational Vision Science \& Technology}, 9(13):32--32,
  December 2020.
\newblock Publisher: The Association for Research in Vision and Ophthalmology.

\bibitem{tramer2017ensemble}
Florian Tram{\`e}r, Alexey Kurakin, Nicolas Papernot, Ian Goodfellow, Dan
  Boneh, and Patrick McDaniel.
\newblock Ensemble adversarial training: Attacks and defenses.
\newblock {\em arXiv preprint arXiv:1705.07204}, 2017.

\bibitem{benabid_deep_2003}
Alim~Louis Benabid.
\newblock Deep brain stimulation for {Parkinson}’s disease.
\newblock {\em Current Opinion in Neurobiology}, 13(6):696--706, December 2003.

\bibitem{bozorgzadeh_neurochemostat_2016}
Bardia Bozorgzadeh, Douglas~R. Schuweiler, Martin~J. Bobak, Paul~A. Garris, and
  Pedram Mohseni.
\newblock Neurochemostat: {A} {Neural} {Interface} {SoC} {With} {Integrated}
  {Chemometrics} for {Closed}-{Loop} {Regulation} of {Brain} {Dopamine}.
\newblock {\em IEEE Transactions on Biomedical Circuits and Systems},
  10(3):654--667, June 2016.
\newblock Conference Name: IEEE Transactions on Biomedical Circuits and
  Systems.

\bibitem{beyeler_pulse2percept_2017}
M.~Beyeler, G.~M. Boynton, I.~Fine, and A.~Rokem.
\newblock pulse2percept: {A} {Python}-based simulation framework for bionic
  vision.
\newblock In K.~Huff, D.~Lippa, D.~Niederhut, and M.~Pacer, editors, {\em
  Proceedings of the 16th {Science} in {Python} {Conference}}, pages 81--88,
  2017.

\bibitem{loshchilov2017decoupled}
Ilya Loshchilov and Frank Hutter.
\newblock Decoupled weight decay regularization.
\newblock {\em arXiv preprint arXiv:1711.05101}, 2017.

\end{thebibliography}



}

\newpage
\section*{Appendix}
\appendix
\counterwithin{figure}{section}
\section{Surrogate Model} \label{app:surrogate}
This section covers specific implementation details about the surrogate model as well as observations on its performance.
\subsection{Implementation Details}
\paragraph{Dataset} To create training data for the surrogate model $\hat{f}_\phi$, we used the phosphene model described in \cite{granley_computational_2021} and implemented in pulse2percept v0.8 \cite{beyeler_pulse2percept_2017}. $50,000$ stimuli were created by first selecting a number of electrodes to stimulate between 1 and 30 randomly chosen electrodes, then randomly selecting an amplitude between 1 and 10 (specified as a multiple of the assumed threshold current) and frequency between 1 and 200 Hz for each electrode. In addition, between 10 and 100 electrodes were chosen to act as ``noise'' electrodes, where either amplitude or frequency was given a nonzero value, but not both.
The purpose of these electrodes was for the surrogate model to learn that both a nonzero amplitude and a nonzero frequency are required to produce a visible percept.
We used an 80-20 train-test split.
As the surrogate model is highly dependent on patient-specific parameters $\phi$, we generated new data and fit a separate surrogate for each of the following $\phi$: ($(\rho, \lambda)\in\{(150,100), (150,1500), (800,100), (800,1500)\}$).

\paragraph{Network Architecture}
The surrogate model $\hat{f}_\phi$ used a fully-connected architecture. The input to the model was a stimulus matrix $\mathbf{s} \in \mathbb{R}_{\geq0}^{n_e \mathrm{x} 3}$, which was identical to the input to $f$. 
The stimulus matrix was split into amplitude and frequency components (pulse duration was not used due to poor model performance), which were fed through a \ac{FC} layer. 
The outputs of both \ac{FC} layers were concatenated and fed through another \ac{FC} layer. Concurrently, the model computed the element-wise product of the amplitude and frequency components and passed it through a separate \ac{FC} layer. The outputs of the previous two layers were then concatenated and fed through a final \ac{FC} layer with output size $49\times49$.

The model was trained for 45 epochs using AdamW \citep{loshchilov2017decoupled} optimizer and MAE loss.

\subsection{Approximating the Forward Model}

The surrogate model was able to accurately approximate the true phosphene model $f$. Table \ref{tab:surrogate} shows MAE over the validation set ($10,000$ percepts) for all 4 trained $\hat{f}_\phi$. Visually, the predicted percepts were nearly identical to the ground truth.

\begin{table}[h]
    \centering
    \caption{Surrogate model performance}
    \label{tab:surrogate}
    \resizebox{\columnwidth}{!}{%
    \begin{tabular}{@{}c|cccc@{}}
    \toprule
    $\phi$ & $\rho=150$ $\lambda=100$ & $\rho=150$ $\lambda=1500$ & $\rho=800$ $\lambda=100$ & $\rho=800$ $\lambda=1500$ \\ \midrule
    MAE    & 0.0119                  & 0.0189                   & 0.0078                  & 0.0115                   \\ \bottomrule
    \end{tabular}
    }
\end{table}

\subsection{Predicted Stimuli}

\begin{figure}[h]
    \centering
    \includegraphics[width=\linewidth]{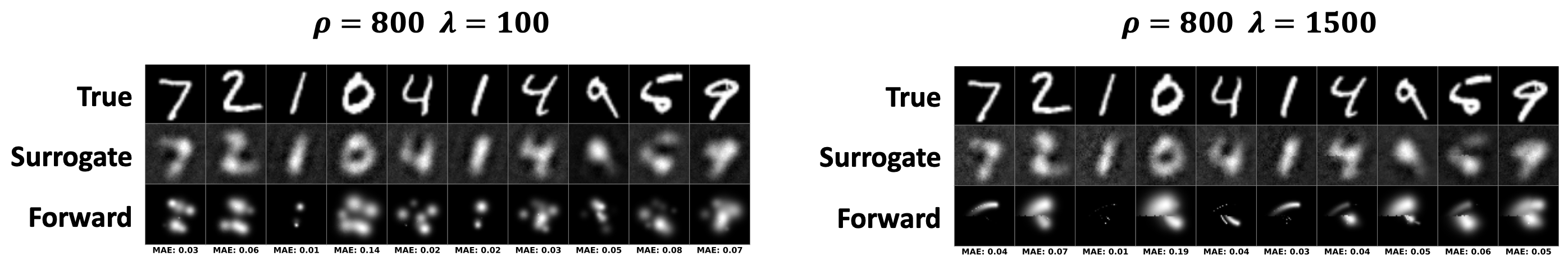}
    \caption{The encoder would often suggest stimuli that lead to drastically different percepts when fed through the surrogate model ($\hat{f}$, middle row) as compared to the true forward model ($f$, bottom row). Examples are shown for $\rho=800$; $\lambda=100$ (\emph{left}) and $\rho=800$ $\lambda=1500$ (\emph{right}).}
    \label{fig:true_vs_surrogate}
\end{figure}

Despite the low surrogate validation error, training with the surrogate model would often result in the encoder suggesting almost adversarial stimuli; that is, stimuli that if fed through the true forward model $f$ would lead to drastically different percepts than if fed through the surrogate model $\hat{f}$ (see Fig.~\ref{fig:true_vs_surrogate}). With these adversarial-like stimuli, the encoder appears to be performing well under the surrogate model, but performs poorly when the same stimuli are input to the true forward model. We identify this as the primary disadvantage of using a surrogate model and resolving this issue remains an open research problem for end-to-end training with surrogate methods.

We noticed several issues caused by the effects of varying stimulus parameters on phosphene appearance. For example, increasing amplitude increases size and brightness, while increasing frequency increases brightness only. We noticed a larger mismatch between the surrogate and the forward model on the extreme ends of the spectrum (\emph{e.g.} very high frequency, low amplitude), resulting in the encoder settling into a minimum that does not exist in the true forward model. It is important to note this disparity appears despite a high training accuracy of the surrogate alone. 
Although these examples are specific to the bionic vision application, we expect surrogate models derived to describe other neuromodulation technologies to suffer from similar limitations.

\newpage
\section{Hyperparameter Selection}\label{app:hyperparameter}
In this section, we detail how \ac{HNA} hyperparameters ($l$, $k$, $\alpha$, and $\beta$) were chosen.

\begin{figure}[ht!]
    \centering
    \includegraphics[width=\linewidth]{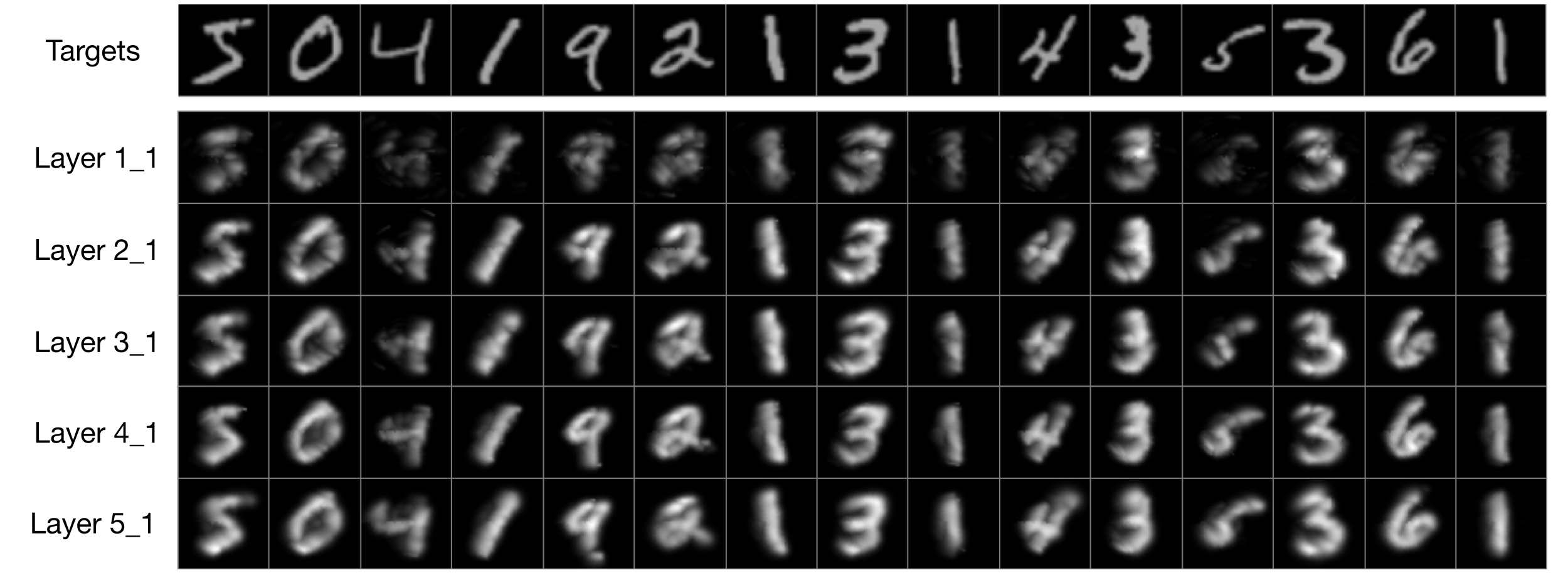}
    \caption{Phosphenes produced by \ac{HNA} encoder with different layers chosen for VGG loss. Layer 5\_1 denotes the first layer within the fifth convolutional block.}
    \label{fig:crossval}
\end{figure}

\paragraph{VGG Loss}
To choose the layer of the VGG network to use for VGG loss ($l$) we performed cross validation across a set of candidate layers. Previous studies \cite{li_universal_2017} have shown that the first layer with each of the 5 convolutional blocks perform well for neural style transfer. Thus, we choose these as our candidate layers. For cross validation, we trained \ac{HNA} for 50 epochs using each candidate layer. The resulting phosphenes are shown in Figure \ref{fig:crossval}. Using earlier layers, the VGG term performs similarly to \ac{MAE}, and phosphenes are disconnected. We chose layer 5\_1 based on its perceived ability to capture high-level perceptual differences between images, although layer 4\_1 also performs similarly.

\paragraph{Laplacian Smoothing}
We chose to use a kernel size 5 for the Laplacian filter used to estimate the second derivative ($k$, Eq. \ref{eq:smooth}). The size of the filter controls the scale on which smoothing is applied (\textit{i.e.}, smaller filters sizes only encourage continuity within a small local region, whereas larger filters encourage continuity within a larger region). Size 5 was chosen because larger filters were observed to over-smooth the image, while smaller filters still led to highly disconnected phosphenes.

\paragraph{Joint Perceptual Metric}
We performed cross validation to find the best values for $\alpha$ and $\beta$. Instead of using one value, we found scheduled weighting to be crucial for performance. The scheduler incrementally increased the weight of the VGG loss ($\beta$) from 0 while simultaneously decreasing the initially high weight on the smoothing constraint ($\alpha$). 
This was motivated by the observation that the VGG loss performed poorly during early iterations when the predicted phosphene was near-random.

Under this scheduled weighting strategy, the loss is dominated early on by the \ac{MAE} and smoothing terms. This encourages the the model to just output reasonable encodings. As training progresses, the predicted phosphenes become higher quality, causing the VGG loss to perform better, and thus the smoothing term is no longer as important. 

Additionally, we found it beneficial to temporarily decrease the learning rate by a factor of 10 for a short 'warm-up' duration following each increase in $\beta$, before resetting to 50\% of the prior learning rate. This results in the learning rate gradually decreasing throughout training by a factor of around 100. Throughout the paper, we use $\alpha=0$ and $\beta=0.00008$ for comparisons of loss values.

\newpage
\section{COCO Dataset} \label{app:coco_dataset}
For the COCO task (Section \ref{sec:coco}), we used subset of images from the MS-COCO dataset \cite{lin_microsoft_2015}. MS-COCO was chosen due to its selection of common household objects relevant to the daily life of prosthesis users, as well as availability of ground-truth segmentation masks. To select the images suitable for prosthetic vision, we filtered out images according to the following criteria:
\begin{itemize}[topsep=0pt, noitemsep, leftmargin=20pt]
    \item[1.] \textbf{Too cluttered.} Any image with greater than 15 total objects was removed. Removed: 15566
    \item[2.] \textbf{Select chosen objects.} Any image that did not have at least 1 object from the selected categories that was larger than 4\% of the total image was removed. Removed: 42289
    \item[3.] \textbf{Too many.} Any image with greater than 5 objects meeting criteria 2 was removed. Removed: 1017
    \item[4.] \textbf{Too dim.} Any objects in the image with average pixel brightness less than 50 were discarded. If this resulted in an image having 0 remaining objects, the image was removed. Removed: 434
\end{itemize}
\begin{figure}[ht!]
    \centering
    \includegraphics[width=\linewidth]{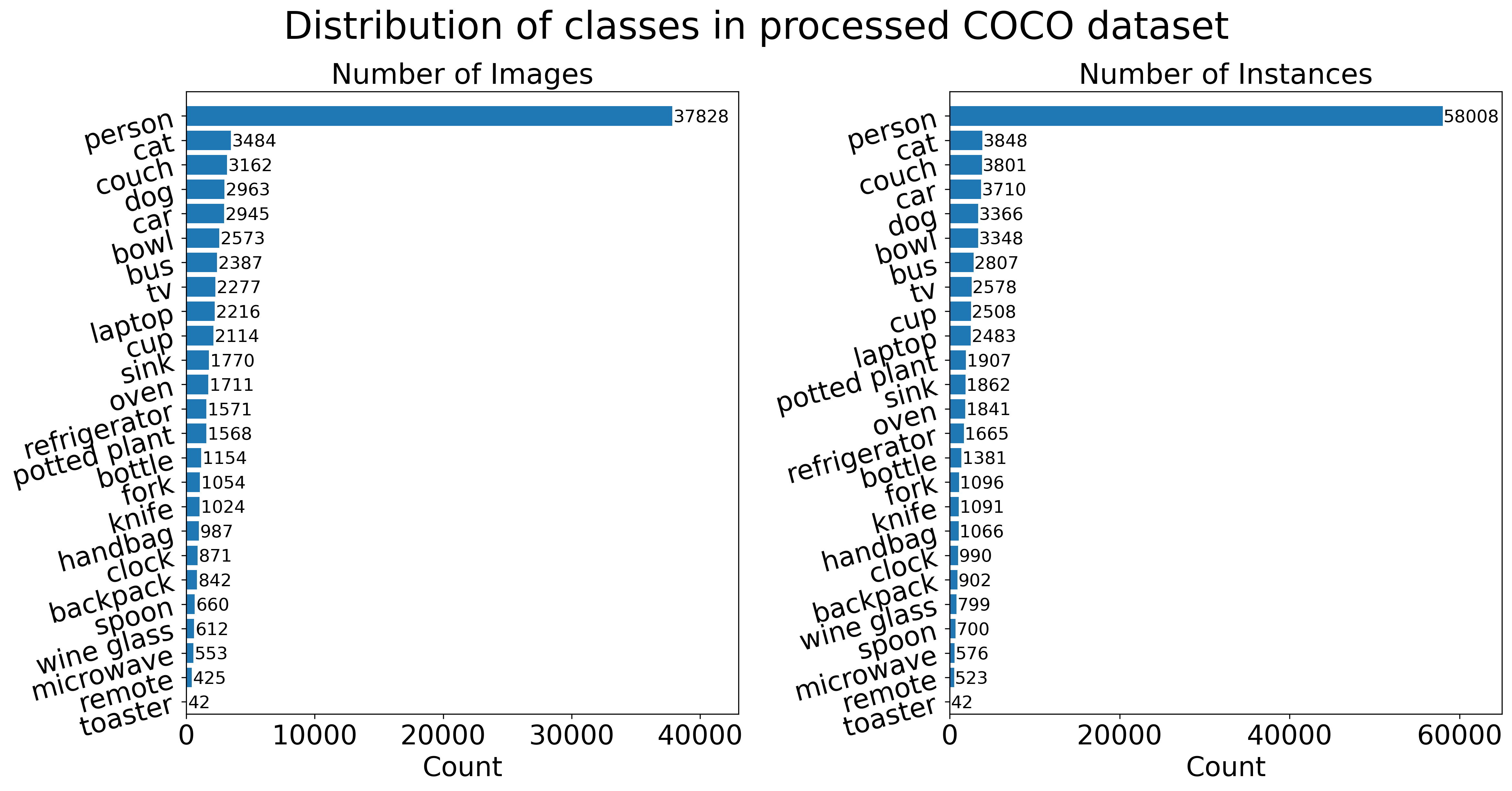}
    \caption{Number of images (\emph{left}) or instances (\emph{right}) of each category in the processed COCO dataset.}
    \label{fig:coco_classes}
\end{figure}

This resulted in a total of 47,532 training images and 11,883 test images (80-20 train-test split). The objects in the remaining images were segmented out using the ground-truth segmentation masks, resized to (49, 49), and converted to grayscale. The distributions of classes used is shown in Figure \ref{fig:coco_classes}.

\newpage
\section{Predicted Stimuli} \label{app:stimuli}
\begin{figure}[h!]
    \centering
    \includegraphics[width=0.9\linewidth]{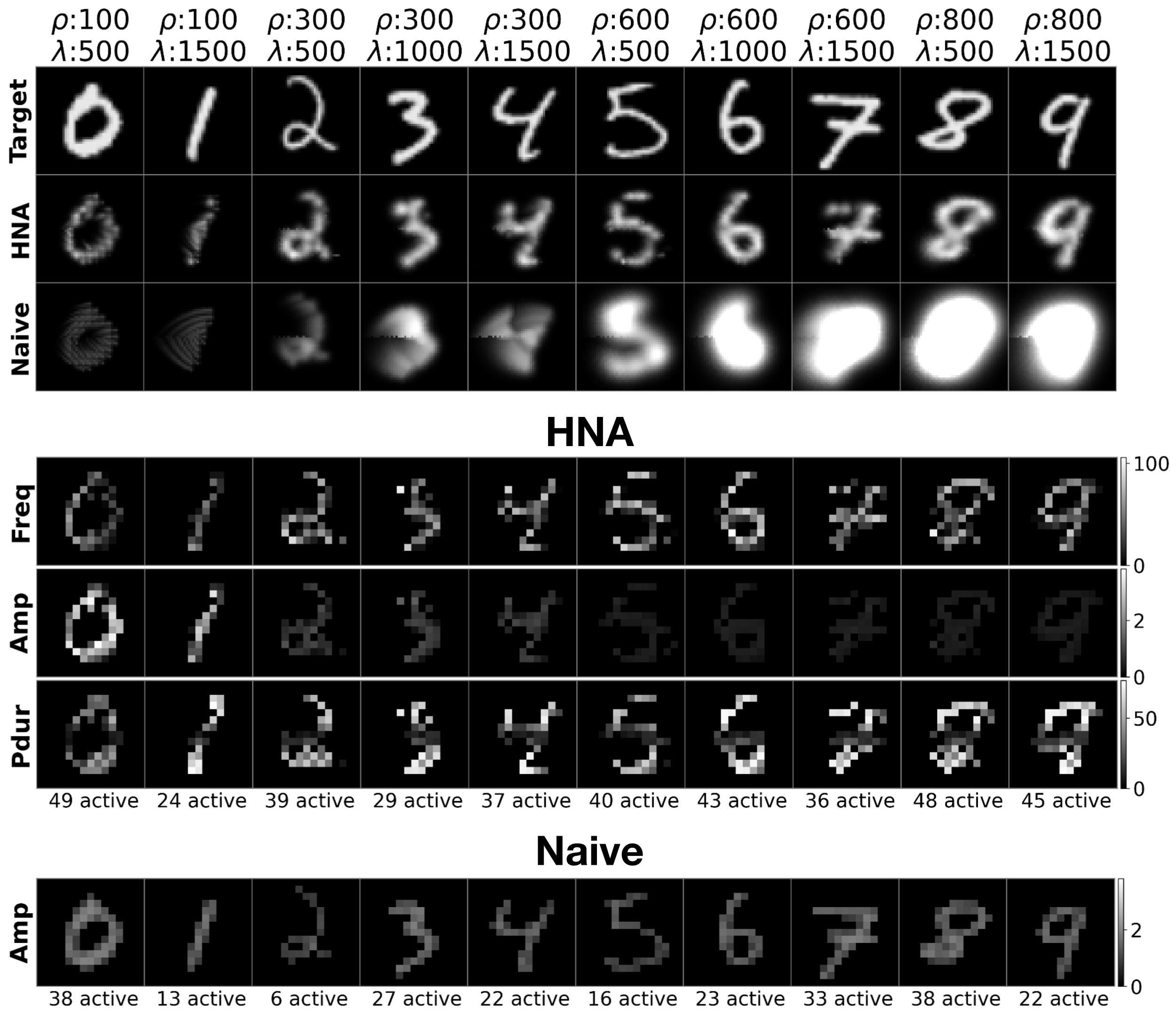}
    \caption{\emph{Top}: Example MNIST target images, and the phosphenes produced by \ac{HNA} and naive encoders, encoded at various $\rho$ and $\lambda$ values. \emph{Center}: The stimuli corresponding to the \ac{HNA} phosphenes. From top to bottom, stimulus frequency (Hz), amplitude (xTh), and pulse duration (ms) are shown. The number of 'active' electrodes stimulated above threshold levels is given below each stimuli. \emph{Bottom}: Stimuli corresponding to the naive phosphenes.}
    \label{fig:stimuli}
\end{figure}

Here, we directly examine the stimuli resulting from \ac{HNA} and naive encoders. Stimuli and their resulting phosphenes for example images from the test set are shown in Figure \ref{fig:stimuli}. The naive encoder produces stimuli with constant frequency (20 Hz) and pulse duration (0.45 ms), which are not shown. 

We make the following observations about the predicted stimuli:
\begin{itemize}[topsep=0pt, noitemsep, leftmargin=20pt]
    \item Both encoders activate electrodes corresponding to the shape of the target image. In naive stimuli, the amplitude directly corresponds to the pixel brightness. In \ac{HNA} stimuli, the distributions of amplitude, frequency, and pulse duration across the electrodes is more complex and harder to characterize, but lead to higher-quality phosphenes.
    \item \ac{HNA} uses amplitudes inversely proportional to $\rho$.
    \item For small $\rho$, \ac{HNA} primarily uses amplitude to control brightness. For large $\rho$, \ac{HNA} primarily uses frequency to modulate brightness, keeping amplitudes low to limit phosphene size. 
    \item \ac{HNA} uses small pulse durations to create lines parallel to the underlying axon \ac{NFB} (\textit{i.e.}, it utilizes the streaked phosphenes to its advantage), and large pulse durations to create lines perpendicular to the underlying \ac{NFB}. In other words, \ac{HNA} was able to exploit application-specific (\emph{i.e.}, neuroanatomical) information that is baked into the forward model.
    \item On average, \ac{HNA} uses more electrodes, larger frequencies and pulse durations, and smaller amplitudes than the naive encoder. A large active electrode count and high pulse durations may not be desirable for some prostheses, due to tissue activation and frame rate limits. We found that it was easy to constrain these parameters using regularization on the output stimuli, at the cost of slightly decreased performance.
\end{itemize}

\newpage
\section{COCO Patient-to-Patient Variations} \label{app:coco_patient}
\begin{figure}[ht!]
    \centering
    \includegraphics[width=\linewidth]{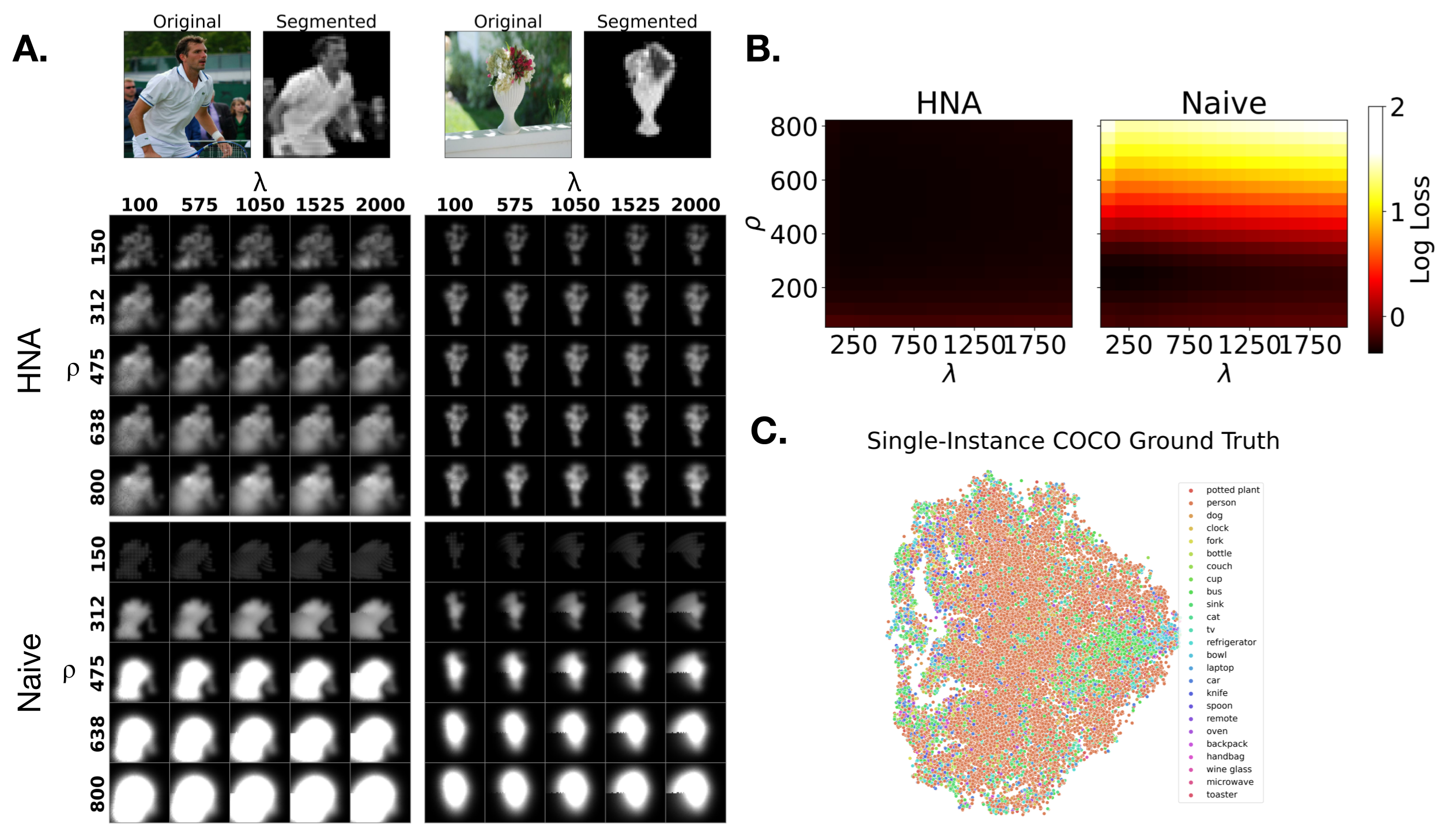}
    \caption{COCO Encoder performance across simulated patients (varying $\rho$ and $\lambda$). \textbf{A:} Phosphenes produced by HNA and Naive encoders of two example images. \textbf{B:} Heatmaps showing the log joint loss across $\rho$ and $\lambda$ for \ac{HNA} and naive encoders. \textbf{C.} Ground-truth COCO images cannot be clustered using T-SNE into groups corresponding to the object types. The clustering was performed on COCO images that only contained one object.}
    \label{fig:coco_patient-to-patient}
\end{figure}

We repeated the analysis presented in Section \ref{sec:patient-to-patient} for the COCO dataset. Figure \ref{fig:coco_patient-to-patient}A shows two example COCO images, encoded by both \ac{HNA} and the naive encoder, across varying $\rho$ and $\lambda$ values. The heatmaps in Figure \ref{fig:coco_patient-to-patient}B show the log of the joint perceptual loss across simulated patients, for both the naive and \ac{HNA} encoders. To measure phosphene consistency, we performed T-SNE clustering on a subset of the COCO images which have only 1 object. Unfortunately, T-SNE clustering of the ground-truth COCO images did not form groups corresponding to the object types (Figure \ref{fig:coco_patient-to-patient}C), suggesting that the representation of object instances vary drastically across COCO images. Therefore, it was not meaningful to repeat the analysis presented in Fig.~\ref{fig:mnist-rho-lam}C.

Similar to the MNIST results presented in Section \ref{sec:patient-to-patient}, \ac{HNA} produced higher-quality representations than the naive encoder, resulting in a lower joint loss for every simulated patient. \ac{HNA} performed consistently well across all simulated patients (Figure \ref{fig:coco_patient-to-patient}B), with a small increase in loss for small $\rho$ ($< 100$). Similar to MNIST, the naive encoder only performs well for patients with a mid-to-low $\rho$ ($\approx 200$) and low $\lambda$.

\newpage
\section{Modeling Other Patient-to-Patient Variations} \label{app:other_modelparams}
\begin{figure}[ht!]
    \centering
    \includegraphics[width=\linewidth]{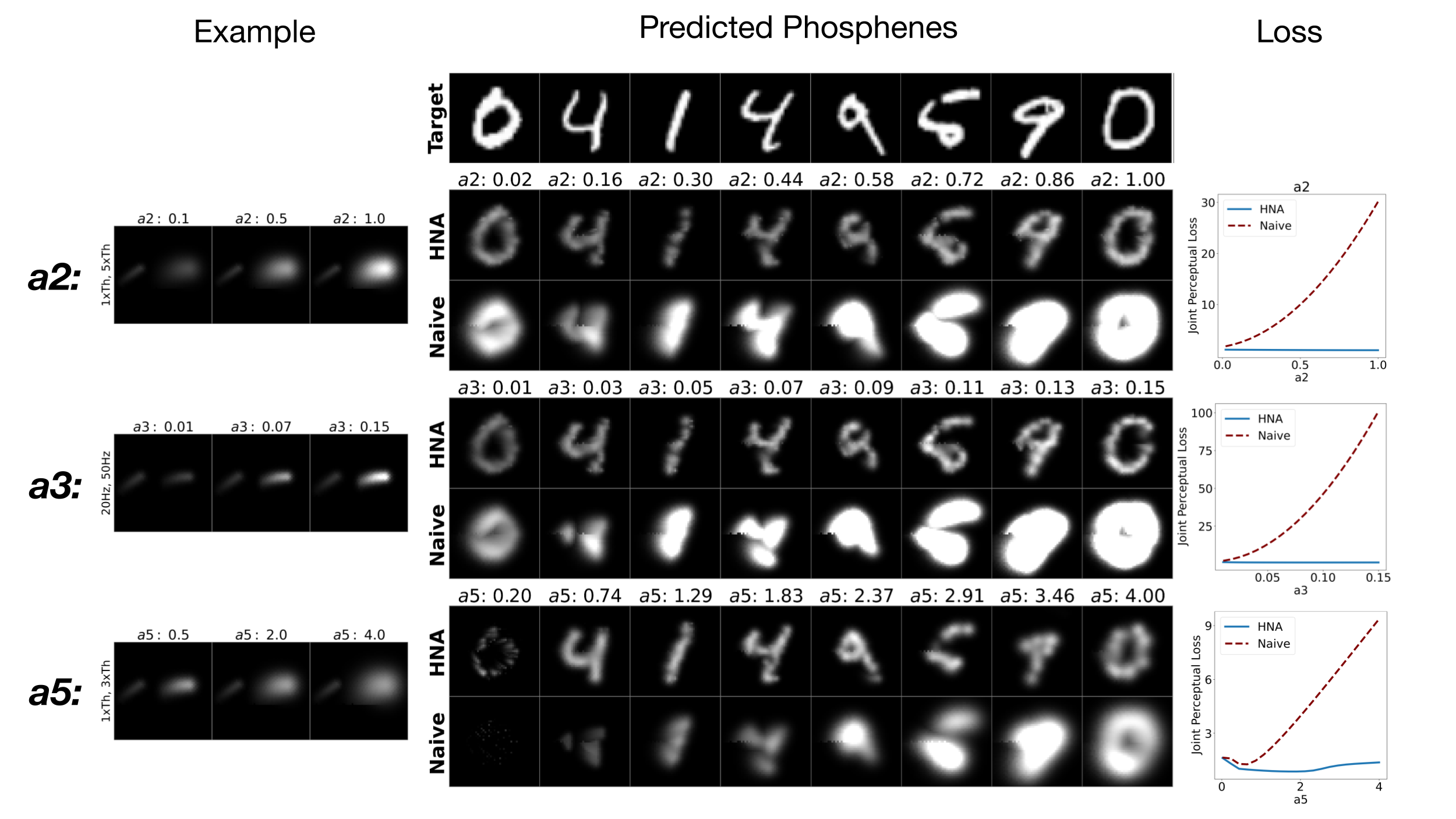}
    \caption{\emph{Left}: Examples of how a2, a3, and a5 affect single-electrode phosphenes. a2 modulates local brightness scaling with increasing amplitude, a3 modulates local brightness scaling with increasing frequency, and a5 modulates local size scaling with increasing amplitude. \emph{Center}: Phosphenes predicted with \ac{HNA} and naive encoders for varying a2, a3, and a5, increasing left to right. \emph{Right}: Plot showing the joint loss across a2, a3, and a5 for \ac{HNA} (solid) and naive encoder (dashed line).}
    \label{fig:modelparams}
\end{figure}

Previously, results were presented across patient-specific parameters $\rho$ and $\lambda$, because these have the greatest impact on phosphene appearance. However, the forward model has a number of other patient-specific parameters, which \ac{HNA} is also able to adapt to. For full details on all parameters of the forward model, see \cite{granley_computational_2021}. Out of the remaining parameters, $a2$, $a3$, and $a5$ are the most impactful on phosphene appearance. $a2$ and $a3$ modulate how much the brightness contribution from each electrode scales with increasing amplitude and frequency, respectively. $a5$ locally scales the global radial current spread $\rho$ based on each electrodes amplitude. Figure \ref{fig:modelparams} (\emph{left}) illustrates the effect of these parameters on phosphene appearance.

Figure \ref{fig:modelparams} compares \ac{HNA} to naive encoder performance across $a2$, $a3$, and $a5$. The ranges for these parameters are based on values empirically observed in retinal prosthesis users \cite{granley_computational_2021}. \ac{HNA} produces relatively consistent phosphenes, and outperforms the naive encoder across all conditions. 

\newpage
\section{Simulating Higher-Resolution Implants} \label{app:high-res}
\begin{figure}[h]
    \centering
    \includegraphics[width=\linewidth]{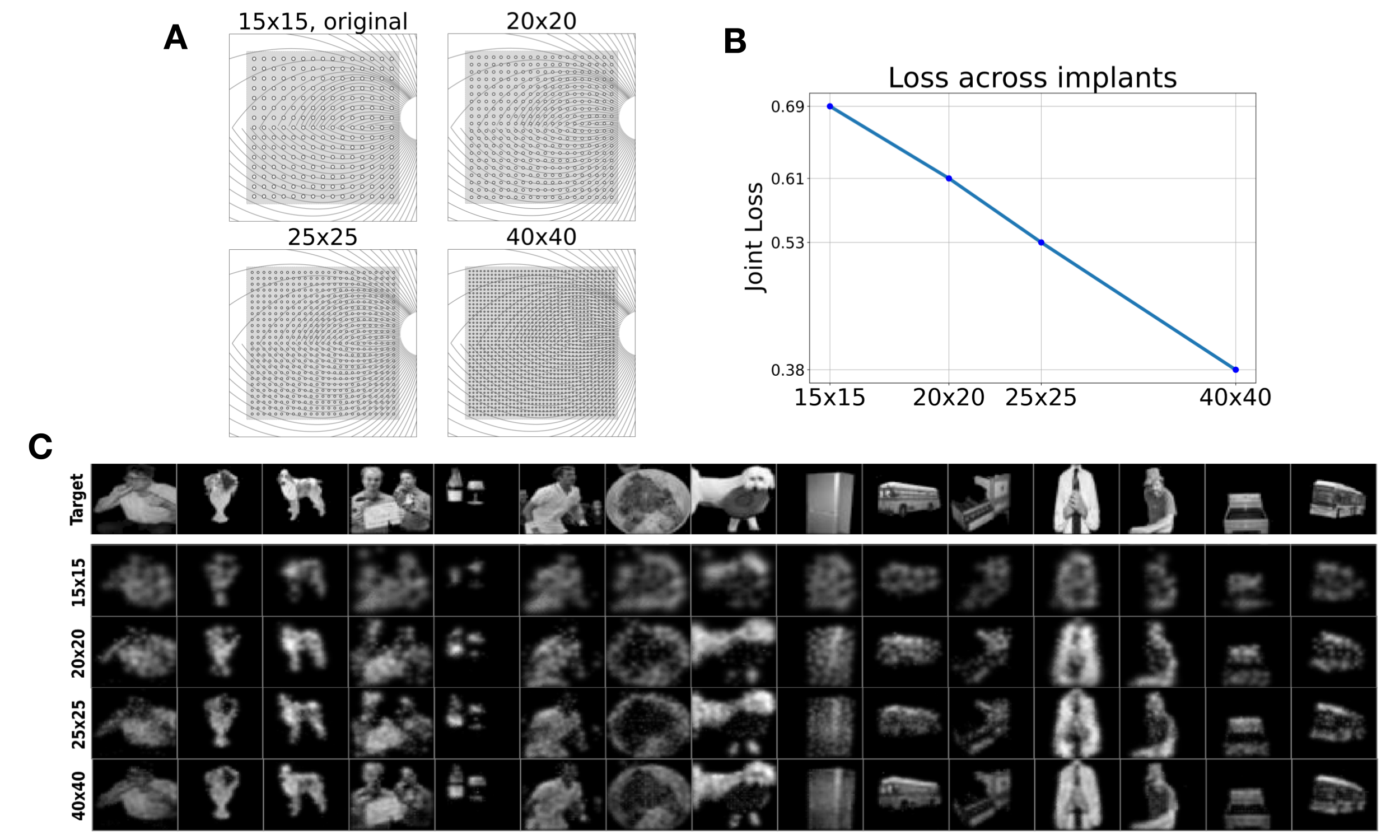}
    \caption{\textbf{A}: The 4 different implants compared. The main text uses the $15 \times 15$ implant. \textbf{B}: The joint perceptual loss of \ac{HNA}s trained on the different implants after 50 epochs. \textbf{C}: Example images showing the reconstructed phosphenes using each implant}
    \label{fig:implants}
\end{figure}

On the COCO task, \ac{HNA} significantly outperformed the naive encoder, but was still unable to capture all of the detail in the images. Two of the main reasons for this are the limited spatial resolution of the implant and the patient-specific distortions from the forward model. Here, we present results from \ac{HNA}s trained on implants of higher resolution, at small $\rho$ and $\lambda$. The chosen implants are illustrated in Figure \ref{fig:implants}A. For a fair comparison, each \ac{HNA} was trained for only 50 epochs. 

Phosphenes resulting from the \ac{HNA} trained on the different implants are shown in Figure \ref{fig:implants}C, and the losses across implants is plotted in Figure \ref{fig:implants}B. As implant resolution increases, the phosphenes look increasingly similar to the ground truth, and small details (e.g. facial details, textures) start to emerge.


Thus, \ac{HNA}s initial failure to capture high-frequency details in the image appears to be an application-specific limitation for visual prostheses more so than a limitation of the \ac{HNA} framework. For visual prostheses, learning to reconstruct the high-frequency features of complex images despite distortions and limited implant resolution remains an open problem.

\newpage
\section{Mis-Specified Patient-Specific Parameters}
\label{app:mis-specified}
Due to noisy or limited patient data, there may be some uncertainty in the measured value of the patient-specific parameters $\phi$. Therefore, we conducted an analysis of the consequences of incorrect patient-specific parameters on the encodings produced by \ac{HNA}. Note that the true patient-specific parameters are not needed during training, so incorrect $\phi$ will only affect evaluation. A 'mismatch' \ac{HNA} model was created, where the forward model decoder used the true patient-specific parameters $\phi$, and the encoder used another set of patient-specific parameters $\phi'$.

In the first experiment, $\phi'$ was sampled from a uniform random distribution (we again focus on only $\rho$ and $\lambda$). The original \ac{HNA} encoder, naive encoder, and mismatch \ac{HNA} encoder with random $\phi'$ were evaluated on the MNIST test set. \ac{HNA} achieved a joint loss of 0.92, the naive encoder had a joint loss of 3.13, and the mismatch \ac{HNA} had a joint loss of 1.35 $\pm$ 0.003 (mean $\pm$ standard deviation across 10 random $\phi'$). Thus, even if the true patient-specific parameters are completely unknown, on average randomly selecting values will still produce higher-quality encodings than the naive method.

\begin{figure}[hb!]
    \centering
    \includegraphics[width=\linewidth]{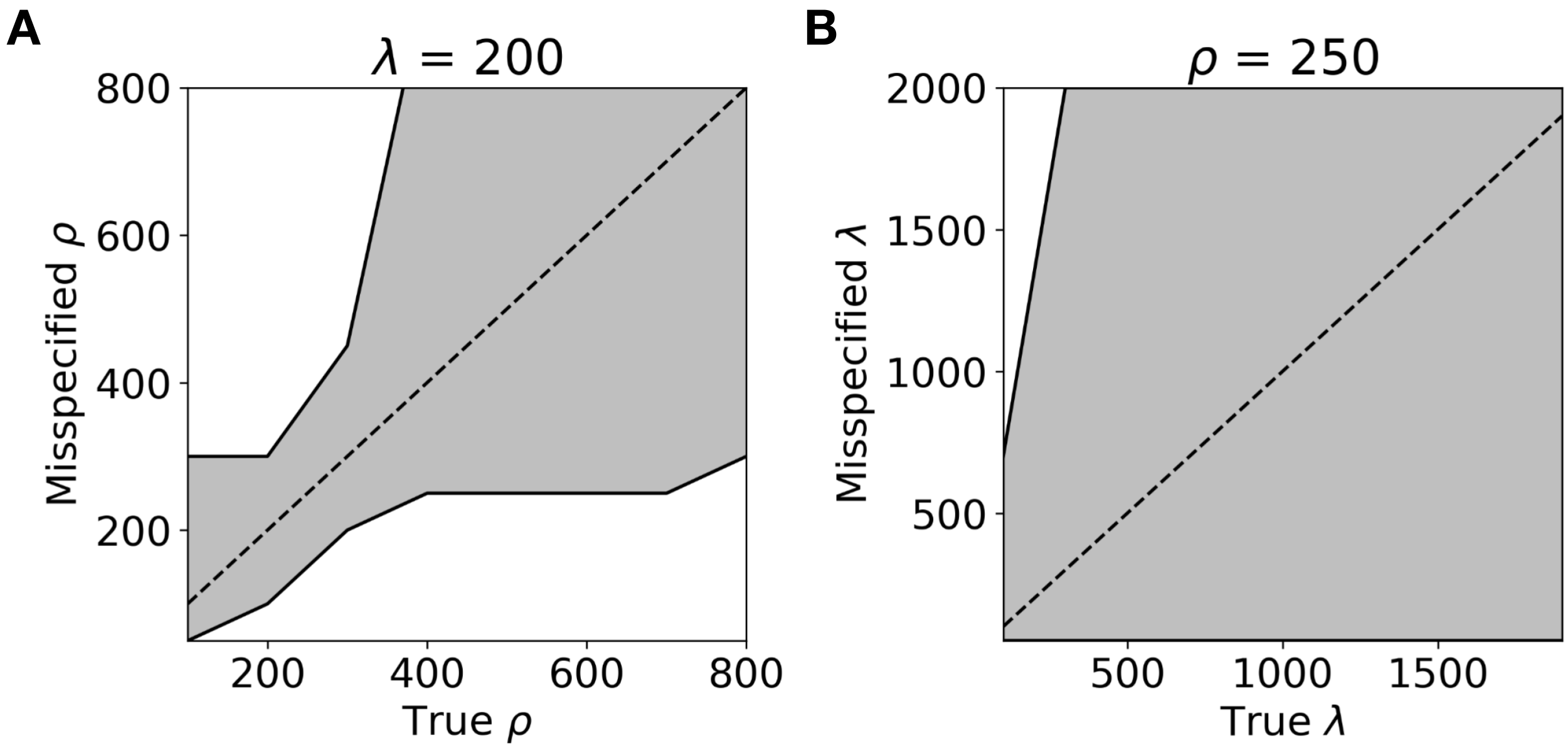}
    \caption{Plots showing mis-specified \ac{HNA} performance relative to the naive encoder for varying $\rho$ (panel \textbf{A}) and $\lambda$ (panel \textbf{B}). The dashed line marks the correctly specified model, and shaded area between the solid lines shows the region where the mis-specified \ac{HNA} outperforms the naive encoder. Note that the naive model's ideal patient was used, with $\lambda$ fixed at 200 and $\rho$ fixed at 250, respectively.}
    \label{fig:misspecified}
\end{figure}

In a second experiment, we analyzed whether there were any configurations ($\phi$ - $\phi'$ combinations) that resulted in a worse encoding than the naive model. For the 90\% of true $\phi$, the mismatch model outperformed the naive model regardless of the chosen $\phi'$. However, the naive model performs best at $\rho=250$ and $\lambda=200$. In Figure \ref{fig:misspecified}A, we hold $\lambda$ constant at $200$ and, for each true $\rho$, plot the ranges of mis-specified $\rho'$ for which the mismatch \ac{HNA} still outperforms the naive. Figure \ref{fig:misspecified}B shows a similar plot for varying $\lambda$, holding $\rho$ constant at $250$. Even for the naive model's ideal patients, \ac{HNA} still outperforms the naive model for a large proportion of mis-specified $\rho$ and $\lambda$. 

\end{document}